\documentclass{article}
\usepackage[accepted]{icml2021}
\usepackage[utf8]{inputenc}
\usepackage{url}
\usepackage{hyperref}
\usepackage[dvipsnames]{xcolor}
\usepackage{multirow}
\usepackage{amsmath,amsfonts,bm,amssymb,amsthm,yfonts}

\def\eqref#1{equation~\ref{#1}}
\def\1{\bm{1}}

\def\rr{{\textnormal{r}}}

\def\vzero{{\bm{0}}}
\def\vone{{\bm{1}}}

\def\vf{{\bm{f}}}

\def\vu{{\bm{u}}}

\def\vw{{\bm{w}}}
\def\vx{{\bm{x}}}

\def\rr{\mathbb{R}}

\DeclareMathAlphabet{\mathsfit}{\encodingdefault}{\sfdefault}{m}{sl}
\SetMathAlphabet{\mathsfit}{bold}{\encodingdefault}{\sfdefault}{bx}{n}

\def\t{\intercal}

\def\eee#1#2{\mathbb{E}_{#1}\left[#2\right]}

\newtheorem{theorem}{Theorem}

\newtheorem{lemma}{Lemma}

\newtheorem{assumption}{Assumption}

\usepackage{listings}
\usepackage{accents}
\usepackage{booktabs} 
\usepackage{threeparttable}

\def\var{\mathbb{V}}
\def\dd{\mathrm{d}}

\def\tot{\mathrm{tot}}

\def\aug{\mathrm{aug}}

\def\vec{\mathrm{vec}}
\def\ee#1{\mathbb{E}\left[#1\right]}

\def\ema{\mathrm{a}}

\def\eee#1#2{\mathbb{E}_{#1}\left[#2\right]}

\usepackage{tabularx,colortbl,xcolor}

\newif\ifcomments
\commentsfalse

\newif\ifarxiv
\arxivfalse
\ifcomments
\usepackage[textsize=small,color=lightgray]{todonotes}
\paperwidth=\dimexpr \paperwidth + 10cm\relax
\paperheight=\dimexpr \paperheight + 5cm\relax
\oddsidemargin=\dimexpr\oddsidemargin + 5cm\relax
\evensidemargin=\dimexpr\evensidemargin + 5cm\relax
\marginparwidth=\dimexpr \marginparwidth + 5cm\relax
\else
\usepackage[disable]{todonotes}
\fi

\def\ourtitle{Understanding Self-Supervised Learning Dynamics without Contrastive Pairs}
\def\pred{p}
\def\vec{\mathrm{vec}}
\def\diag{\mathrm{diag}}

\def\ourmethod{\textbf{DirectPred}} 

\icmltitlerunning{\ourtitle}

\begin{document}

\twocolumn[
\icmltitle{\ourtitle}

\icmlsetsymbol{equal}{*}

\begin{icmlauthorlist}
\icmlauthor{Yuandong Tian}{fair}
\icmlauthor{Xinlei Chen}{fair}
\icmlauthor{Surya Ganguli}{fair,stanford}
\end{icmlauthorlist}

\icmlaffiliation{fair}{Facebook AI Research}
\icmlaffiliation{stanford}{Stanford University}

\icmlcorrespondingauthor{Yuandong Tian}{yuandong@fb.com}

\icmlkeywords{Machine Learning, ICML}

\vskip 0.3in
]

\printAffiliationsAndNotice{}

%In the case of linear predictors, our approach outperforms gradient based training of the predictor by $5\%$. 
\vspace{-0.03in}
\begin{abstract}
While contrastive approaches of self-supervised learning (SSL) learn representations by minimizing the distance between two augmented views of the same data point (positive pairs) and maximizing views from different data points (negative pairs), recent \emph{non-contrastive} SSL (e.g., BYOL and SimSiam) show remarkable performance {\it without} negative pairs, with an extra learnable predictor and a stop-gradient operation. A fundamental question arises: why do these methods not collapse into trivial representations? We answer this question via a simple theoretical study and propose a novel approach, \ourmethod{}, that \emph{directly} sets the linear predictor based on the statistics of its inputs, without gradient training. On ImageNet, it performs comparably with more complex two-layer non-linear predictors that employ BatchNorm and outperforms a linear predictor by $2.5\%$ in 300-epoch training (and $5\%$ in 60-epoch). \ourmethod{} is motivated by our theoretical study of the nonlinear learning dynamics of non-contrastive SSL in simple linear networks. Our study yields conceptual insights into how non-contrastive SSL methods learn, how they avoid representational collapse, and how multiple factors, like predictor networks, stop-gradients, exponential moving averages, and weight decay all come into play. Our simple theory recapitulates the results of real-world ablation studies in both STL-10 and ImageNet. Code is released\footnote{\url{https://github.com/facebookresearch/luckmatters/tree/master/ssl}}.
\end{abstract}

\vspace{-0.1in}
\section{Introduction}
Self-supervised learning (SSL) has emerged as a powerful method for learning useful representations without requiring expensive target labels~\cite{bert}. Many state-of-the-art SSL methods in computer vision employ the principle of contrastive learning \cite{oord2018representation,tian2019contrastive,he2020momentum,simclr,bachman2019learning} whereby the hidden representations of two augmented views of the same object (positive pairs) are brought closer together, while those of different objects (negative pairs) are encouraged to be further apart. Minimizing differences between positive pairs encourages modeling invariances, while contrasting negative pairs is thought to be required to prevent representational collapse (i.e., mapping all data to the same representation).

\begin{figure}[t]
    \centering
    \includegraphics[width=0.45\textwidth]{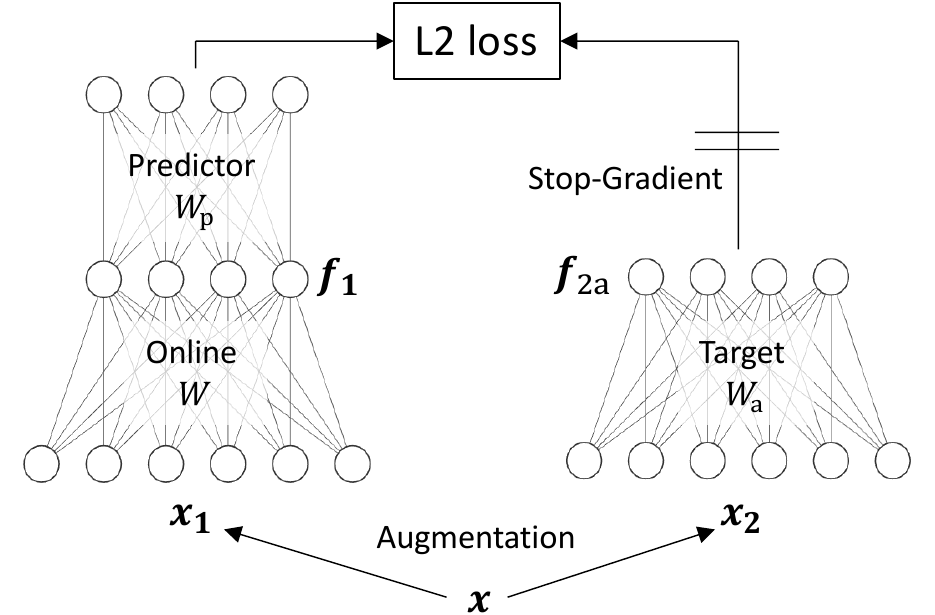}
    \vspace{-0.1in}
    \caption{Two-layer setting with a linear, bias-free predictor.}
    \label{fig:two-layer-setting}
\end{figure}

However, some recent SSL work, notably BYOL \cite{byol} and SimSiam \cite{chen2020exploring}, have shown the remarkable capacity to learn powerful representations using only positive pairs, {\it without} ever contrasting negative pairs. These methods employ a dual pair of Siamese networks \cite{Bromley1994-mk} (Fig. \ref{fig:two-layer-setting}): the representation of two views are trained to match, one obtained by the composition of an online and predictor network, and the other by a target network. The target network is {\it not} trained via gradient descent; and either employs a direct copy of the online network (e.g., SimSiam \cite{chen2020exploring}), or a momentum encoder that slowly follows the online network in a delayed fashion through an exponential moving average (EMA) (e.g., MoCo~\cite{he2020momentum,chen2020improved} and BYOL~\cite{byol}). Compared to contrastive learning, these non-contrastive SSL methods do not require large batch size (e.g., 4096 in SimCLR~\cite{simclr}) or memory queue (e.g., MoCo~\cite{he2020momentum,chen2020improved}) to provide negative pairs. Therefore, they are generally more efficient and conceptually simple while maintaining state-of-the-art performance.

Since the entire procedure in non-contrastive SSL encourages the online+predictor network and the target network to become similar to each other, this overall scheme raises several fundamental unsolved theoretical questions. Why/how does it avoid collapsed representations? What is the nature of the learned representations? How do multiple design choices and hyperparameters interact nonlinearly in the learning dynamics? While there are interesting theoretical studies of contrastive SSL ~\cite{Arora2019-xw,lee2020predicting,tosh2020contrastive}, any theoretical understanding of the nonlinear learning dynamics of non-contrastive SSL remains open.

In this paper, we make a first attempt to analyze the behavior of non-contrastive SSL training and the empirical effects of multiple hyperparameters, including \textbf{(1)} Exponential Moving Average (EMA) or momentum encoder, \textbf{(2)} Higher relative learning rate ($\alpha_\pred$) of the predictor, and \textbf{(3)} Weight decay $\eta$. We explain all these empirical findings with an exceedingly simple theory based on analyzing the nonlinear learning dynamics of simple linear networks. Note that deep linear networks have provided a useful tractable theoretical model of nonconvex loss landscapes \cite{kawaguchi2016deep,du2019width,laurent2018deep} and nonlinear learning dynamics ~\cite{saxe2013exact,Saxe2019-eg,Lampinen2018-sl,arora2018optimization} in these landscapes, yielding insights like dynamical isometry \cite{saxe2013exact,Pennington2017-za,Pennington2018-fy} that lead to improved training of nonlinear deep networks. Despite the simplicity of our theory, it can still predict how various hyperparameter choices affect performance in an extensive set of real-world ablation studies. Moreover, the simplicity also enables us to provide conceptual and analytic insights into {\it why} performance patterns vary the way they do. Specifically, our theory accounts for the following diverse empirical findings:

\textbf{Essential part of non-contrastive SSL.} The existence of the predictor and stop-gradient is absolutely essential. Removing either of them leads to representational collapse in BYOL and SimSiam.

\textbf{EMA}. While the original BYOL needs EMA to work, they later confirmed that EMA is not necessary (i.e., the online and target networks can be identical) if a higher $\alpha_\pred$ is used. This is also confirmed with SimSiam, as long as the predictor is updated more often or has larger learning rate (or larger $\alpha_\pred$). However, the performance is slightly lower.   

\textbf{Predictor Optimality and Relative learning rate $\alpha_\pred$}. Both BYOL and SimSiam suggest that the predictor should always be optimal, in the sense of always achieving minimal $\ell_2$ error in predicting the target network's outputs from the online network's outputs. This optimality conjecture was motivated by observed superior performance when the predictor had large learning rates and/or was allowed more frequent updates than the rest of the network. However~\cite{chen2020exploring} also showed that if the predictor is updated too often, then performance drops, which questions the importance of an always optimal predictor as a key requirement for learning good representations. 

\textbf{Weight Decay}. Table 15 in BYOL~\cite{byol} indicates that no weight decay may lead to unstable results. A recent blogpost~\cite{byol-bn-blog} also mentions using weight decay leads to stable learning in BYOL.

Finally, motivated by our theoretical analysis, we propose a new method \ourmethod{} that directly sets the predictor weights based on principal components analysis of the predictor's input, thereby avoiding complicated predictor dynamics and initialization issues. We show that this simple \ourmethod{} method nevertheless yields comparable performance in CIFAR-10 and outperforms gradient training of the linear predictor by $+5\%$ Top-1 accuracy in linear evaluation protocol on both STL-10 and ImageNet (60 epochs). On the standard ImageNet benchmark (300 epochs), \ourmethod{} achieves $72.4\%/91.0\%$ Top-1/Top-5, $2.5\%$ higher than BYOL with linear predictor ($69.9\%/89.6\%$) and comparable with default BYOL setting with 2-layer predictor ($72.5\%/90.8\%$). 

\begin{table}[t]
    \centering
    \small
    \setlength{\tabcolsep}{1.2pt}
    \begin{tabular}{c||c|c|c|c}
                    & \multicolumn{4}{c}{Plug-in frequency (every $N$ minibatches)} \\
                    & 1 & 2 & 3 & 5 \\ 
         \hline
         EMA        & $40.67{\pm}0.50$ & $35.29{\pm}2.49$ & $34.60{\pm}0.98$ & $35.63{\pm}2.66$ \\
         no EMA     & $39.45{\pm}1.26$ & $34.01{\pm}1.54$ & $34.58{\pm}2.93$ & $32.22{\pm}2.94$ 
    \end{tabular}
    \caption{\small Simply plugging in the ``optimal solution'' to the linear predictor shows poor performance after 100 BYOL epochs (Top-1 accuracy in STL-10~\cite{coates2011analysis} downstream classification task). The optimal solution is obtained by solving (with regularization) $W_\pred \ee{\vf\vf^\t} = \frac{1}{2}(\ee{\vf_\ema \vf^\t} + \ee{\vf \vf_\ema^\t})$, in which the two expectations is estimated with exponential moving average. In comparison, with gradient descent, BYOL with a single linear layer predictor can reach 74\%-75\% Top-1 in STL-10 after 100 epochs. Unless explicitly stated, in all our experiments, we use ResNet-18~\cite{resnet} as the backbone network for CIFAR-10/STL-10 experiments and SGD as the optimizer with learning rate $\alpha = 0.03$, momentum $0.9$, weight decay $\bar\eta=0.0004$ and EMA parameter $\gamma_\ema=0.996$. Each setting is repeated 5 times.\label{tab:optimal-do-not-work}}
\end{table}

\section{Two-layer linear model} 
\label{sec:two-layer-model}
To obtain analytic and conceptual insights into non-contrastive SSL we analyze a simple, \emph{bias-free} linear BYOL model where the online, target and predictor networks are specified by the weight matrices $W\in \rr^{n_2\times n_1}$, $W_p \in \rr^{n_2\times n_2}$ and $W_\ema \in \rr^{n_2\times n_1}$ respectively (Fig.~\ref{fig:two-layer-setting}). Let $\vx \in \mathbb{R}^{n_1}$ be a data point drawn from the data distribution $p(\vx)$ and let 
$\vx_1$ and $\vx_2$ be two augmented views of
$\vx$: $\vx_1, \vx_2\sim p_\aug(\cdot|\vx)$ where $p_\aug(\cdot | \vx)$ is the augmentation distribution. 
In practice such data augmentations correspond to random crops, blurs or color distortions of images~\cite{simclr}. 
Let $\vf_1 = W\vx_1 \in\rr^{n_2}$ be the online representation of view $1$, and $\vf_{2\ema}=W_\ema\vx_2\in\rr^{n_2}$ be the target representation of view $2$. In BYOL, the learning dynamics of $W$ and $W_p$ are obtained by minimizing
\begin{equation}
    J(W, W_\pred) := \frac{1}{2}\eee{\vx_1, \vx_2}{\|W_\pred \vf_1 - \mathrm{StopGrad}(\vf_{2\ema})\|^2_2}, \label{eq:objective}
\end{equation}
while the dynamics of $W_a$ is obtained differently, via an exponential moving average (EMA) of $W$. We will analyze this combined dynamics for $W$, $W_p$ and $W_a$, in the presence of additional weight decay, in the limit of large batch sizes and small discrete time learning rates. This limit can be well approximated by the gradient flow (see Supplementary Material (SM) for all derivations):
\begin{lemma}
\label{lemma:gradient-rule}
BYOL learning dynamics following Eqn.~\ref{eq:objective}:
\begin{eqnarray}
    \dot W_\pred \!\!&\!\! = \!\!&\!\! \alpha_\pred\left(- W_p W(X+X') + W_\ema X\right) W^\t - \eta W_\pred \label{eq:grad-wp} \\
    \dot W       \!\!&\!\! = \!\!&\!\! W^\t_p \left(- W_\pred W (X+X') + W_\ema X\right) - \eta W \label{eq:grad-w} \\
    \dot W_\ema  \!\!&\!\! = \!\!&\!\! \beta (- W_\ema + W) \label{eq:ema} 
\end{eqnarray}
\end{lemma}

Here, $X := \ee{\bar\vx\bar\vx^\t}$ where $\bar\vx(\vx) := \eee{\vx'\sim p_\aug(\cdot|\vx)}{\vx'}$ is the average augmented view of a data point $\vx$ and  $X' := \eee{\vx}{\var_{\vx'|\vx}[\vx']}$ is the covariance matrix $\var_{\vx'|\vx}[\vx']$ of augmented views $\vx'$ conditioned on $\vx$, subsequently averaged over the data $\vx$. Note that $\alpha_p$ and $\beta$ reflect \emph{multiplicative learning rate ratios} between the predictor and target networks relative to the online network. Finally, the terms involving $\eta$ reflect weight decay. 

As a gradient flow formulation, the learning rate $\alpha$ does not appear in Lemma~\ref{lemma:gradient-rule}. In the actual finite time update, the learning rate for $W_\pred$ is $\alpha\alpha_\pred$, the EMA rate is $\alpha\beta = 1 - \gamma_\ema$, where $\gamma_\ema$ is the usual EMA parameter (e.g,. BYOL uses $0.996$), and the weight decay for actual training is $\bar\eta:=\alpha\eta$.

We note that since SimSiam is an ablation of BYOL that removes the EMA computation, the underlying dynamics of SimSiam can also be obtained from Lemma~\ref{lemma:gradient-rule} simply by setting $W_a=W$, inserting this relation into Eqn.~\ref{eq:grad-wp} and Eqn.~\ref{eq:grad-w}, and ignoring Eqn.~\ref{eq:ema}. Importantly, the stop-gradient on the target branch is still there. 

Overall Eqns.~\ref{eq:grad-wp}-\ref{eq:ema} constitute our starting point for analyzing the combined roles of relative learning rates $\alpha_p$ and $\beta$, weight decay rate $\eta$ and various ablations in determining the performance of both BYOL and SimSiam.  

We first derive two very general results (see SM).

\begin{theorem}[Weight decay promotes balancing of the predictor and online networks.]
\label{thm:invariance}
Completely independent of the particular dynamics of $W_a$ in Eqn.~\ref{eq:ema}, the update rules (Eqn.~\ref{eq:grad-wp} and Eqn.~\ref{eq:grad-w}) possess the invariance 
\begin{equation}
    W(t)W^\t(t) = \alpha^{-1}_p W_p^\t(t) W_p(t) + e^{-2\eta t}C, 
\end{equation}
where $C$ is a symmetric matrix that depends only on the initialization of $W$ and $W_p$.
\end{theorem}
This theorem implies that for both BYOL and SimSiam, there exists a ``balancing'' that ensures that any matching between the online and target representations will not be attributable solely to the predictor weights, rendering the online weights useless. Instead what the predictor learns, the online network will also learn, which is important as the online network's representations are what is used for downstream tasks.  We note that similar weight balancing dynamics has been discovered in multi-layer linear networks and matrix factorization~\cite{arora2018optimization,du2018algorithmic}. Our results generalize this to SSL dynamics. Second, a nonzero weight decay could help remove the extra constant $C$ due to initialization, further balancing the predictor and online network weights and possibly leading to better performance on downstream tasks (Tbl.~\ref{tab:no-weight-decay}). 

\begin{table}[]
    \centering
    \small
    \setlength{\tabcolsep}{1pt}
    \begin{tabular}{c|c|c|c}
     EMA + no-bias & EMA + bias & no EMA + no-bias & no EMA + bias \\
     \hline 
     $70.62{\pm}1.05$& $70.99{\pm}1.01$ & $71.36{\pm}0.44$& $71.37{\pm}0.77$
    \end{tabular}
    \vspace{-0.1in}
    \caption{\small Top-1 accuracy of BYOL on STL-10 under linear evaluation protocol, trained for 100 epochs with no weight decay ($\eta = 0$) and $\alpha_\pred=1$. It is worse than the baseline ($74.51 {\pm}0.47$ without predictor bias) when the weight decay is set to be $\eta = 0.0004$. ``No-bias'' means the linear predictor does not have a bias term.}
    \label{tab:no-weight-decay}
\end{table}

\begin{theorem}[The stop-gradient signal is essential for success.]
\label{thm:nostop-faiedl}
With $W_\ema=W$ (SimSiam case), removing the stop-gradient signal yields a gradient update for $W$ given by positive semi-definite (PSD) matrix $H(t) := X'\otimes(W_p^\t W_p+I_{n_2}) + X\otimes\tilde W_p^\t \tilde W_p + \eta I_{n_1n_2}$ (here $\tilde W_p := W_p - I_{n_2}$ and $\otimes$ is the Kronecker product):  
\begin{equation}
    \frac{\dd}{\dd t}\vec(W) = - H(t) \vec(W).  
\end{equation}
If the minimal eigenvalue $\lambda_{\min}(H(t))$ over time is bounded below, $\inf_{t\ge 0} \lambda_{\min}(H(t)) \ge \lambda_0 > 0$, then $W(t)\rightarrow 0$. 
\end{theorem}
Thus we have proven analytically in this simple setting that removing the stop-gradient leads to representational collapse, as observed in more complex settings in SimSiam~\cite{chen2020exploring}. Similarly, with $W_a=W$ and no predictor ($W_p=I_{n_2}$), then the dynamics Eqn.~\ref{eq:grad-w} also reduces to a similar form and $W(t)\rightarrow 0$ (see SM). 

\begin{figure*}
    \centering
    \includegraphics[width=0.95\textwidth]{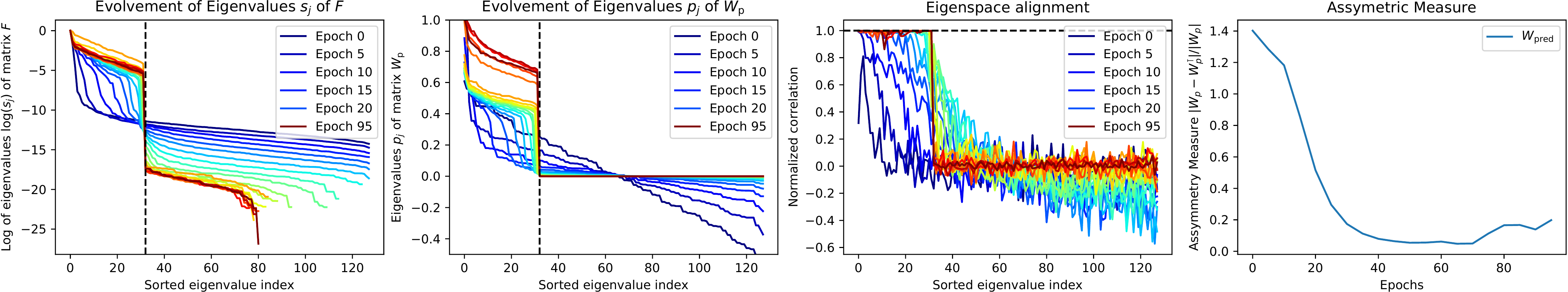}
    \includegraphics[width=0.95\textwidth]{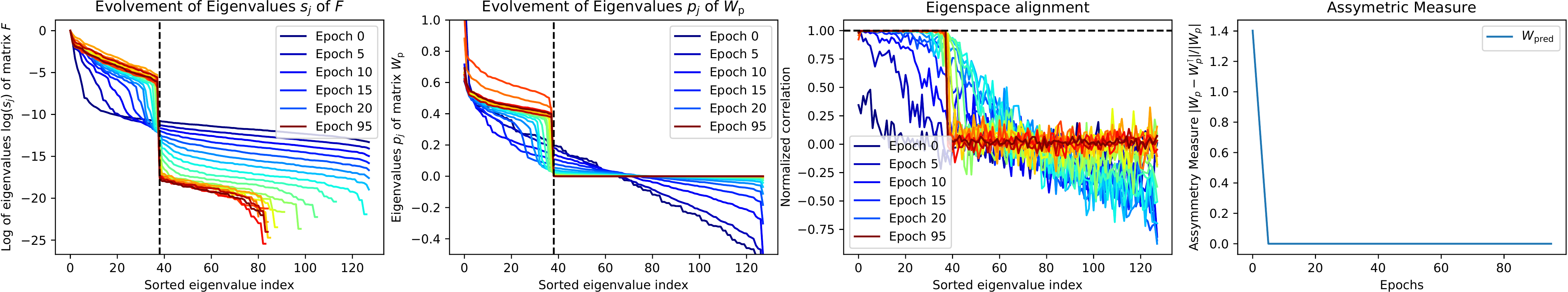}
    \vspace{-0.15in}
    \caption{\small Training BYOL in STL-10 for 100 epochs with EMA. \textbf{Top row}: No symmetric regularization imposed on $W_p$, \textbf{Bottom row}: symmetric regularization on $W_\pred$. From left to right: \textbf{(1)} Evolvement of eigenvalues for $F$. Since $F$ is PSD and its eigenvalue $s_j$ varies across scales, we plot $\log(s_i)$. We could see some eigenvalues are growing while others are shrinking to zero over training. \textbf{(2)} Similar ``step-function'' behaviors for the predictor $W_\pred$. Its negative eigenvalues shrinks towards zero and leading eigenvalues becomes larger. \textbf{(3)} The eigenspace of $F$ and $W_\pred$ gradually align with each other (Theorem~\ref{thm:alignment-eigenspace}). For each eigenvector $\vu_j$ of $F$, we compute cosine angle (normalized correlation) between $\vu_j$ and $W_\pred \vu_j$ to measure alignment. \textbf{(4)} $W_\pred$ gradually becomes symmetric and PSD during training.}    \label{fig:eigen-space-alignment}
\end{figure*}

\section{How multiple factors affect learning dynamics}
\label{sec:assumption}
The learning dynamics in Eqns.~\ref{eq:grad-wp}-\ref{eq:ema} constitute a set of high dimensional coupled nonlinear differential equations that can be difficult to solve analytically in general. Therefore, to obtain analytic insights into the functional roles of the relative learning rates $\alpha_\pred$ and $\beta$ and weight decay $\eta$, we make a series of simplifying assumptions.  Intriguingly, under these simplifying assumptions we obtain a rich set of analytic predictions, which we then test experimentally in more realistic scenarios. We find, nicely, that these predictions still qualitatively hold {\it even when} our simplifying assumptions required for obtaining analytic results do not. 

\begin{assumption}[Proportional EMA]
\label{assumption:ema}
We first reduce the dimensionality of the dynamics in Eqns.~\ref{eq:grad-wp}-\ref{eq:ema} by enforcing that the target network $W_a$ undergoes EMA but is forced to always be proportional to the online network via the relation $W_\ema(t) = \tau(t) W(t)$.  Inserting this relation into the EMA dynamics in Eqn.~\ref{eq:ema} yields $\dot \tau W + \tau \dot W = \beta(1-\tau)W$.   
\end{assumption}
Thus we obtain a reduced dynamics for $W$, $W_p$ and $\tau$. By not enforcing the stronger SimSiam constraint that $W_a = W$, we can still model EMA dynamics. Intuitively, $\tau = \tau(t)$ is a dynamic parameter that depends on how quickly $W=W(t)$ grows over time. If $W$ is constant, then $\dot W=0$ and $\tau$ stabilizes to $1$. On the other hand, if $W$ grows rapidly, then $\tau$ becomes small. While Assumption~\ref{assumption:ema} is a simplification, as we shall see, it still reveals interesting verifiable predictions about the functional role of EMA. 

\begin{assumption}[Isotropic data and augmentation]
\label{assumption:isotropic}
We assume the data distribution $p(\vx)$ has zero mean and identity covariance, while the augmentation distribution  $p_\aug(\cdot|\vx)$ has mean $\vx$ and covariance $\sigma^2 I$. This simplifies the dynamics in Eqns.~\ref{eq:grad-wp}-\ref{eq:ema} by reducing the augmentation averaged data covariance to $X = I$ and the data averaged augmentation covariance to $X'=\sigma^2 I$.
\end{assumption}

Many previous studies of deep learning dynamics made simplifying isotropic assumptions about data~\cite{tian2017analytical,brutzkus2017globally,du2018gradient,bartlett2018gradient,safran2018spurious}. Since our fundamental goal is to obtain the first analytic understanding of the dynamics of non-contrastive SSL methods, it is useful to first achieve this in the simplest possible isotropic setting. Interestingly, we will find that our final conclusions generalize to non-isotropic real world settings. 

\begin{assumption}[Symmetric predictor]
\label{assumption:symmetricpredictor}
We enforce symmetry in $W_p$ by initializing it to be a symmetric matrix, and then symmetrizing the flow for $W_p$ in Eqn.~\ref{eq:grad-wp} (see SM). 
\end{assumption}

This symmetry assumption was motivated by both fixed point analysis and empirical findings. First, the fixed point of Eqn.~\ref{eq:grad-wp} under Assumption~\ref{assumption:ema} and~\ref{assumption:isotropic} and $\eta > 0$ is always a symmetric matrix and in numerical simulation the asymmetric part $W_p - W_p^\t$ eventually vanishes (See Appendix for the proof and numerical simulations). % Note that while the fixed point is symmetric, under the unconstrained dynamics $W_p$ might still exhibit asymmetric transients even if initialized symmetrically. 
Moreover, during BYOL training without a symmetry constraint on the predictor, $W_\pred$ gradually moves towards symmetry (Fig.~\ref{fig:eigen-space-alignment}).
%Note $W_\pred$ need not approach exact symmetry since Assumption~\ref{assumption:ema}-\ref{assumption:isotropic} may not hold in practice.

%HSec.~\ref{sec:fixed-point}

Second, a set of experiments reveal that whether the predictor is symmetric or not has a dramatic effect in terms of both performance and interaction with EMA. In our STL-10 experiment, enforcing symmetric $W_\pred$ in the presence of EMA \emph{improves} performance on downstream tasks (Tbl.~\ref{tab:symmetric-no-EMA-does-not-work}). In contrast, in the absence of EMA, a symmetric $W_\pred$ fails while an asymmetric $W_\pred$ works reasonably well. Similar behavior holds on ImageNet: a symmetric one layer linear predictor $W_\pred$ in SimSiam (i.e. without EMA) achieves performance no better than random guessing (Top-1/5: $0.1\%/0.5\%$), while an asymmetric $W_\pred$ achieves a Top-1/5 accuracy of $68.1\%/88.2\%$.  Our theory will explain this as well as show how to obtain good performance with a symmetric predictor without EMA by increasing its relative learning rate $\alpha_p$. 

\begin{table}[t]
    \centering
    \small
    \setlength{\tabcolsep}{0.2pt}
    \begin{tabular}{c||c|c||c|c}
                 & \multicolumn{2}{c||}{No predictor bias} & \multicolumn{2}{c}{With predictor bias} \\
                 &  sym $W_\pred$ & regular $W_\pred$ & sym $W_\pred$ & regular $W_\pred$ \\
     \hline
     \multicolumn{5}{c}{\emph{One-layer linear predictor}} \\
     \hline
     EMA    & $75.09{\pm}0.48$  & $74.51{\pm}0.47$ & $74.52{\pm}0.29$ & $74.16{\pm}0.33$\\
     no EMA & {\color{red} $\mathbf{36.62{\pm}1.85}$ } & $72.85{\pm}0.16$ & {\color{red} $\mathbf{36.04{\pm}2.74}$ } & $72.13{\pm}0.53$ \\ 
     \hline
     \multicolumn{5}{c}{\emph{Two-layer predictor with BatchNorm and ReLU}} \\
     \hline 
    EMA & $71.58{\pm}6.46$& $78.85{\pm}0.25$& $77.64{\pm}0.41$& $78.53{\pm}0.34$\\
    no EMA & {\color{red} $\mathbf{35.59{\pm}2.10}$} & $65.98{\pm}0.71$& {\color{red} $\mathbf{41.92{\pm}4.25}$} & $65.59{\pm}0.66$
    \end{tabular}
    \caption{\small The effect of symmetrization of $W_\pred$ on downstream classification task (BYOL Top-1 on STL-10). Symmetric $W_\pred$ leads to slightly better performance compared to regular $W_\pred$ in the presence of EMA. On the other hand, without EMA, symmetric $W_\pred$ crashes. Same effects happen in two-layer predictor with BatchNorm and ReLU as well. Weight decay $\bar\eta=0.0004$ and $\alpha_p = 1$.} 
    \label{tab:symmetric-no-EMA-does-not-work}
\end{table}

\subsection{Dynamical alignment of eigenspaces between the predictor and its input correlation matrix}
\label{sec:eigen-alignment}
Under the three assumptions stated above, we analyze the coupled dynamics of $F:=WXW^\t$ and $W_\pred$. Note that $F$ is the \emph{correlation matrix} of the outputs of the online network which also serve as inputs to the predictor. By Assumption~\ref{assumption:isotropic}, $\ee{\vx} = \vzero$ and $F$ is also the covariance matrix. We find $F$ and $W_p$ obey the following dynamics (see SM):
\begin{eqnarray}
    \dot W_\pred &=& -\frac{\alpha_\pred}{2}(1+\sigma^2)\{W_\pred, F\} + \alpha_\pred\tau F - \eta W_\pred \label{eq:symmetric} \\
    \dot F &=& -(1+\sigma^2) \{W_\pred^2, F\}+ \tau\{W_\pred,F\} - 2\eta F \nonumber
\end{eqnarray}

This dynamics reveals that the eigenspace of $W_\pred$ will gradually align with that of $F$ under certain conditions (see SM for derivation): 

\begin{theorem}[Eigenspace alignment]
\label{thm:alignment-eigenspace}
Under Eqn.~\ref{eq:symmetric}, the \emph{commutator} $[F,W_\pred] := F W_p - W_p F$ satisfies:
\begin{equation}
    \frac{\dd}{\dd t} [F,W_\pred] = -[F,W_\pred] K - K [F,W_\pred] 
\end{equation}
where 
\begin{equation}
K(t) = (1+\sigma^2)\left[\frac{\alpha_\pred}{2}F(t) + W_p^2(t) - \frac{\tau}{1+\sigma^2} W_p(t)\right] + \frac{3}{2}\eta I \label{eq:K}
\end{equation}
If $\inf_{t\ge 0}\lambda_{\min}[K(t)] = \lambda_0 > 0$, then the commutator
\begin{equation}
    \|[F(t),W_\pred(t)]\|_F \le e^{-2\lambda_0 t} \|[F(0),W_\pred(0)]\|_F \rightarrow 0
\end{equation}
For symmetric $W_\pred$, when $W_\pred$ and $F$ commute they can be simultaneously diagonalized. Thus this shows that the eigenspace of $W_p$ gradually aligns with that of $F$. 
\end{theorem}

To test this prediction, we performed extensive experiments showing that training BYOL using ResNet-18 on STL-10 yields eigenspace alignment, as demonstrated in Fig.~\ref{fig:eigen-space-alignment}. 

Now if the eigenspaces of $W_p$ and $F$ do align, we can obtain fully decoupled dynamics. Let the columns of the matrix $U$ be the common eigenvectors, so that $W_p = U\Lambda_{W_p} U^\t$ where $\Lambda_{W_p} = \diag[p_1, p_2, \ldots, p_d]$, $F=U\Lambda_F U^\t$ where $\Lambda_F = \diag[s_1, s_2, \ldots, s_d]$. For each mode $j$, we have (see SM for derivation):
\begin{eqnarray}
    \dot p_j &\!=\!& \alpha_\pred s_j \left[\tau - (1+\sigma^2)p_j\right] - \eta p_j \label{eq:p} \\
    \dot s_j &\!=\!& 2p_js_j\left[\tau -(1+\sigma^2)p_j\right] - 2\eta s_j \label{eq:s}\\
    s_j\dot \tau &\!=\!& \beta (1-\tau)s_j - \tau \dot s_j / 2. \label{eq:tau}
\end{eqnarray}
This decoupled dynamics constitutes a dramatically simplified set of $3$ dimensional nonlinear dynamical systems for BYOL learning, and two dimensional nonlinear systems (obtained by constraining $\tau=1$) for SimSiam. As expected, each mode's dynamics is equivalent to the $3$ dimensional dynamics obtained by setting $n_1=n_2=1$ in Eqns.~\ref{eq:grad-wp}-\ref{eq:ema} and making the replacements $W^2=s_j$, $W_\pred = p_j$, and $W_a/W = \tau$ (see SM).  Thus the decoupled dynamics in Eqns~\ref{eq:p}-~\ref{eq:tau} reduce to the scalar case of BYOL dynamics in Eqns.~\ref{eq:grad-wp}-\ref{eq:ema} after a change of variables and the condition in Thm.~\ref{thm:alignment-eigenspace} reveals when this decoupled regime is reachable. 

\textbf{Non-symmetric $W_p$}. When Assumption~\ref{assumption:symmetricpredictor} is absent, the analysis is much more convoluted. One possible way is to decompose $W_p = A + B$ where $A=A^\t$ is symmetric and $B=-B^\t$ is skew-symmetric. We leave it for future work. 

\subsection{Analysis of decoupled dynamics}
\label{sec:decoupled-dynamics}
The simplified three (two) dimensional dynamics of BYOL (SimSiam) yields significant insights. 
First, there is clearly a collapsed fixed point at $p_j(t) = s_j(t) = 0$ and $\tau$ taking any value. We wish to understand conditions under which $p_j$ and $s_j$ can avoid this collapsed fixed point and grow from small random initial conditions. Since $s_j$ is an eigenvalue of $WW^\t$, we are particularly interested in conditions under which $s_j$ achieves large final values, corresponding to a non-collapsed online network, that are moreover sensitive to the statistics of the data, governed by $\sigma^2$.  

\begin{figure}[t]
    \centering
    \includegraphics[width=0.49\textwidth]{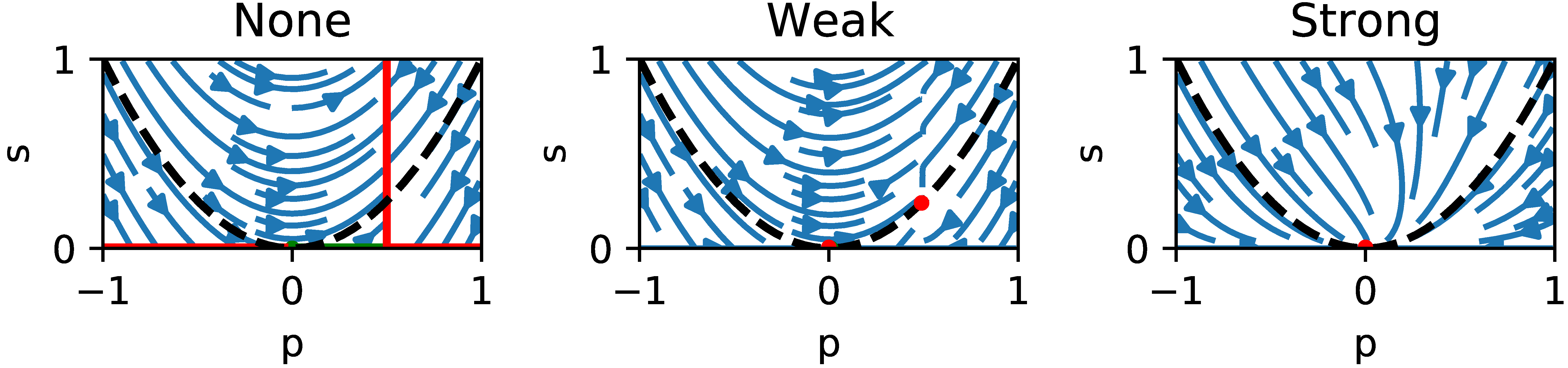}
    \vspace{-0.3in}
    \caption{\small State space dynamics in Eqns.~\ref{eq:p} and \ref{eq:s} for no ($\eta=0$) weak ($\eta=0.01$) and strong ($\eta=1$) weight decay at fixed $\tau=1$ and $\alpha_p=1$.  Red (green) points indicate stable (unstable) fixed points, blue curves indicate flow lines, and the dashed black curve indicates the parabola $s_j=p_j^2/\alpha_p$.}
    \label{fig:statespace}
\end{figure}

\textbf{Exact integral}. First, an important observation, similar to Theorem~\ref{thm:invariance}, is that the dynamics possesses an exact integral of motion, obtained by multiplying Eqn.~\ref{eq:p} by $2\alpha_\pred^{-1} p_j$, subtracting, Eqn.~\ref{eq:s} and integrating over time yielding 
\begin{equation}
    s_j(t) = \alpha_{\pred}^{-1} p_j^2(t) + e^{-2\eta t}c_j \label{eq:s-p-relation}
\end{equation}
where $c_j = \alpha_\pred^{-1} p_j^2(0) - s_j(0)$ is fixed by initial conditions. In absence of weight decay ($\eta=0$), this integral reveals that the initial condition encoded in $c_j$ is never forgotten and the dynamics of $p_j$ and $s_j$ are confined to parabolas of the form $s_j(t)=p_j^2(t) + c_j$, as can be seen by the blue flow lines in Fig.~\ref{fig:statespace}(left). With weight decay ($\eta > 0$) over time the initial condition is forgotten and the dynamics approaches the invariant parabola $s_j = \alpha_{\pred}^{-1} p_j^2$ as can been seen by the approach of the blue flow lines to the black dashed parabola in Fig.~\ref{fig:statespace} right and middle. We discuss these two cases in turn.  First we note that in both cases,  since the EMA computation is often  very slow \cite{byol}, corresponding to small $\beta$, the dynamics of $\tau$ in Eqn.~\ref{eq:tau} is slow relative to that of $p_j$ and $s_j$. Therefore to understand the combined dynamics, we can search for the fixed points that $p_j$ and $s_j$ will rapidly approach at fixed $\tau$. Over time $\tau$ will then either slowly approach $1$ (BYOL) or be always equal to $1$ (SimSiam), and $s_j$ and $p_j$ will follow their $\tau$-dependent fixed points. 

\newcounter{obscount}

\newcommand{\obs}{%
    \stepcounter{obscount}%
    (\underline{Obs\#\theobscount})%
}

\def\obsnum#1{\##1}

\textbf{No weight decay.} When $\eta=0$, Eqns.~\ref{eq:p} and ~\ref{eq:s} at a fixed value of $\tau$ yield a branch of collapsed fixed points given by $s_j=0$ and $p_j$ taking any value, and a branch of non-collapsed fixed points, with $p_j = \tau/(1+\sigma^2)$ and $s_j$ taking any value (horizontal and vertical red/green lines in Fig.~\ref{fig:statespace},left). A sufficient criterion on initial conditions to avoid the collapsed branch is $s_j(0) > p^2_j(0)/\alpha_\pred$ corresponding to lying above the dashed black parabola in Fig.~\ref{fig:statespace},left. This restricted initial condition reveals why a fast predictor (large $\alpha_p$) is advantageous \obs: larger $\alpha_p$ leads to a smaller basin of attraction of the collapsed branch by flattening the dashed parabola. Indeed both BYOL and SimSiam have noted that a fast predictor can help avoid collapse. On the other hand, $\alpha_p$ cannot be infinitely large \obs: since $s_j(+\infty) = s_j(0) + \alpha^{-1}_\pred (p^2_j(+\infty) - p_j^2(0))$, very large $\alpha_\pred$ implies that $s_j$, the final value of the online network characterizing the learned representation, does not grow even if $p_j$ does. This is consistent with results which show that optimizing the predictor too often doesn't work in SimSiam~\cite{chen2020exploring}, and directly setting an ``optimal'' predictor fails as well (Tbl.~\ref{tab:optimal-do-not-work}). The online network needs to grow along with the predictor and that cannot happen if the predictor is too fast.  

\textbf{Advantage of weight decay.} In the non-collapsed branch of fixed points without weight decay (vertical red line in Fig.~\ref{fig:statespace},left), the predictor $p_j$ takes the exact value $\tau/(1+\sigma^2)$, which models the invariance to augmentation correctly: a large data augmentation variance $\sigma^2$ should lead to a small magnitude of the learned representation. Ideally, we want $s_j$ to have the same property. With weight decay $\eta > 0$ in Eqn.~\ref{eq:s-p-relation}, memory of the initial condition $c_j$ fades away, yielding convergence to some point on the invariant parabola $s_j = \alpha_{\pred}^{-1} p_j^2$. \obs: Therefore, by tying the online network to the predictor, weight decay allows $s_j$ to also model invariance to augmentations correctly if the predictor does, regardless of the random initial condition $c_j$. 

\begin{table}[]
    \centering
    \small
    \setlength{\tabcolsep}{2pt}
    \begin{tabular}{c||c|c}
         & Positive effects & Negative effects \\
    \hline
    Relative predictor lr $\alpha_p$ & \obsnum1,\obsnum6  & \obsnum2 \\ 
    Weight decay $\eta$ & \obsnum3,\obsnum7 & \obsnum4,\obsnum5 \\ 
    EMA $\beta$ & \obsnum8 & \obsnum9,\obsnum{10} \\
    \end{tabular}
    \vspace{-0.15in}
    \caption{\small Summarization of positive/negative effects of various hyperparameter choices (EMA $\beta$, relative predictor learning rate $\alpha_\pred$ and weight decay $\eta$). ``\#1'' means (\underline{Obs\#1}) in the text.}
    \label{tab:summarization-parameters}
\end{table}

\paragraph{Dynamics on the invariant parabola.} Because weight decay forces convergence to the invariant parabola $s_j = \alpha^{-1}_p p_j^2$, we next focus on dynamics along this parabola (i.e. $c_j=0$ in Eqn.~\ref{eq:s-p-relation}). In this case, Eqn.~\ref{eq:tau} has a solution: 
\begin{equation}
    \tau(t) = p^{-1}_j(t)\beta e^{-\beta t} \int_0^t p_j(t')e^{\beta t'} \dd t,
\end{equation}
with initial condition $\tau(0)=0$. Inserting the invariant $s_j = \alpha_\pred^{-1}p^2_j$ into Eqn.~\ref{eq:p}, the dynamics of $p_j$ is given by:
\begin{equation}
    \dot p_j = p_j^2\left[\tau(t)-(1+\sigma^2)p_j\right] - \eta p_j. \label{eq:beta-dyn}
\end{equation}
We first analyze the fixed points where $\dot p_j = 0$ at fixed $\tau$.
\begin{figure}
    \centering
    \includegraphics[width=0.45\textwidth]{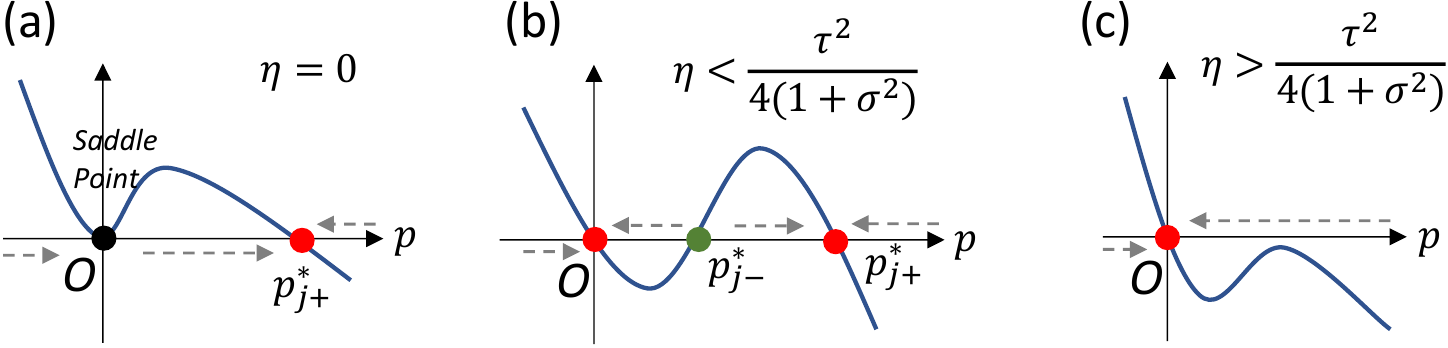}
    \vspace{-0.15in}
    \caption{\small Fixed point of $\dot p_j = -p_j(p_j - p^*_{j-})(p_j - p^*_{j+})$ (see Eqn.~\ref{eq:beta-dyn}). Stable fixed points are in {\color{red} red}, unstable in {\color{ForestGreen} green} and saddle in \textbf{black}. When the weight decay $\eta = 0$, the trivial solution $p_j=0$ is a saddle. When $\eta > 0$, the trivial solution becomes stable near to the origin and initial $p_j$ needs to be large enough to converge to the stable non-collapsed solution $p^*_{j+}$.}
    \label{fig:stationary}
    \vspace{-0.2in}
\end{figure}
When the weight decay $0 < \eta \le \frac{\tau^2}{4(1+\sigma^2)}$,  $p_j$ has has three fixed points (Fig.~\ref{fig:stationary}(b)):
\begin{equation}
    p_{j\pm}^* = \frac{\tau {\pm}\sqrt{\tau^2 - 4\eta(1+\sigma^2)}}{2(1+\sigma^2)} > 0,\quad p^*_{j0} = 0 \nonumber
    \label{eq:p-stationary}
\end{equation}
where both $p^*_{j0}$ and $p^*_{j+}$ are stable and $p^*_{j-}$ is unstable, as shown in Fig.~\ref{fig:stationary}(b). The basin of attraction of the collapsed fixed point $p^*_{j0}=0$ is $p_j < p^*_{j-}$ while the basin of attraction of the useful non-collapsed fixed point $p^*_{j+}$ is $p_j > p^*_{j-}$, yielding an important constraint on initial conditions to avoid collapse. Note that $p^*_{j-}$ is a decreasing function of $\tau$ and increasing function of $\eta$ (see SM). This means that with larger $\eta$, $p^*_{j-}$ moves right and the basin of collapse expands \obs. When $\eta > \frac{\tau^2}{4(1+\sigma^2)}$ there is only one stable fixed point $p^*_{j0} = 0$ (Fig.~\ref{fig:stationary}(c)). Under such strong weight decay collapse is unavoidable \obs.

\begin{figure*}
    \centering
    \includegraphics[width=0.9\textwidth]{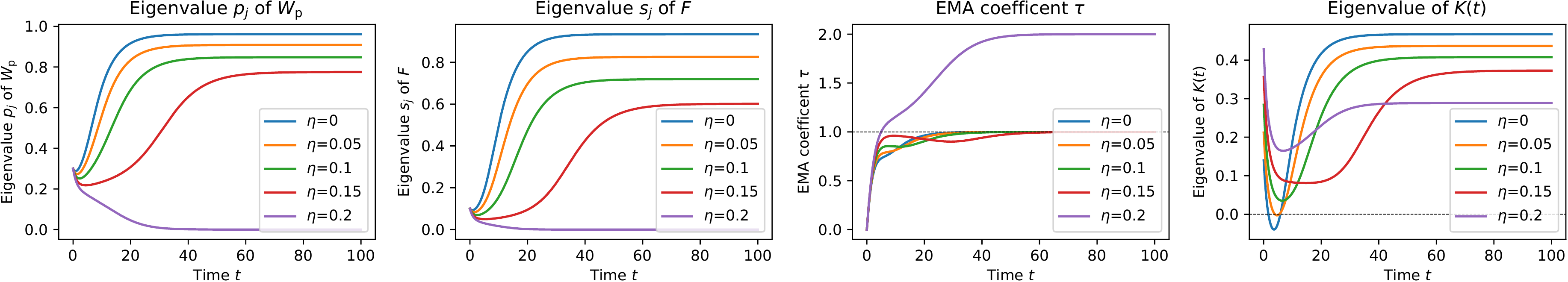}
    \includegraphics[width=0.9\textwidth]{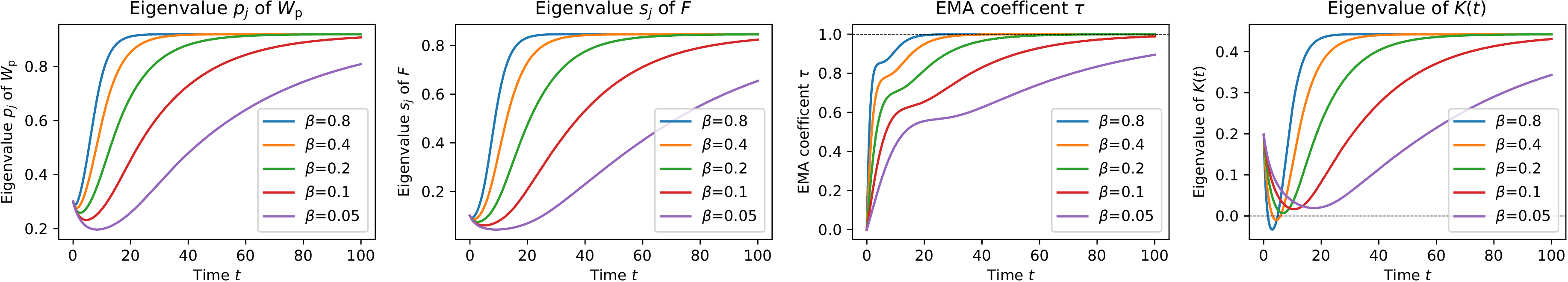}
    \vspace{-0.1in}
    \caption{\small The role played by weight decay $\eta$ and EMA $\beta$ when applying symmetric regularization on $W_\pred$ on synthetic experiments simulating decoupled dynamics (Eqn.~\ref{eq:p}-\ref{eq:tau}). The learning rate $\alpha = 0.01$. Both terms boost the eigenvalue of $K(t)$ to above $0$ so that eigen space alignment could happen (Theorem~\ref{thm:alignment-eigenspace}), but also come with different trade-offs. Here $\beta = 0.4$ so that $\alpha\beta = 0.004 = 1 - \gamma_\ema$ where $\gamma_\ema = 0.996$ as in BYOL.  
    \textbf{Top row (Weight Decay $\eta$)}: A large $\eta$ boost the eigenvalue of $K(t)$ up, but substantially decreases the final converging eigenvalues $p_j$ and $s_j$ (i.e., the final features are not salient), or even drags them to zero (no training happens). \textbf{Bottom row (EMA $\beta$)}. A small EMA $\beta$ also boost the eigenvalue of $K(t)$, but the training converges much slower. Here $\eta = 0.04$ so that $\eta\alpha$ equals to the weight decay ($\bar\eta=0.0004$) in our STL-10 experiments.}
    \label{fig:ablation-eta-beta}
\end{figure*}

We now discuss the dynamics. First we define the quantity $\Delta_j := p_j[\tau - (1+\sigma^2)p_j] -\eta$, which must satisfy \emph{two criteria}.  Note that Eqn.~\ref{eq:beta-dyn} can be written as $\dot p_j = p_j \Delta_j$, so $\Delta_j$ must at some point be positive to drive $p_j(t)$ to any positive non-collapsed fixed point $p^*_{j+}$.  Second, for eigenspace alignment in Theorem~\ref{thm:alignment-eigenspace} to \emph{remain} stable (even if the alignment has already happened), $K(t)$ must be positive definite (PD) in Eqn.~\ref{eq:K}. Using the eigen-space alignment conditions and the invariance $s_j = \alpha^{-1}_\pred p_j^2$, the positive definite condition on $K(t)$ can be written as
\vspace{-0.1in}
\begin{equation}
        \Delta_j < \frac12\left[\alpha_\pred(1+\sigma^2)s_j + \eta\right].  \label{eq:second-condition} 
\end{equation}
%\vspace{-0.1in}
This criterion and the criterion $\Delta_j>0$ yield interesting insights into the roles of various hyperparameters choices. 

First \obs{}, larger predictor learning rate $\alpha_\pred$ can play an advantageous role by loosening the upper bound in Eqn.~\ref{eq:second-condition}, making it easier to satisfy. Second \obs{}, increasing $\eta$ also has the same effect.  

\textbf{Role of EMA}. %\sg{this paragraph is hard to understand} 
Without EMA, $\tau\equiv 1$ and (Eqn.~\ref{eq:second-condition}) may not hold initially when $p_j$ is small. The reason is $\Delta_j$ is to leading order linear in $p_j$ when $\tau=1$ while the right hand side is to leading order $s_j \sim p_j^2$, so the left hand side has a larger contribution from $p_j$ than the right. 

EMA resolves this as follows. When the training begins, $s_j$ is often quite small, and $\tau$ remains small since $W$ changes rapidly. When $p_j$ grows to the fixed point $p^*_{j+} \sim \tau / (1+\sigma^2)$, the growth of $s_j$ stops, making $\tau$ \emph{larger}. This in turns sets a higher fixed point goal for $p_j$. This process continues until the feature is stabilized and $\tau = 1$ (Fig.~\ref{fig:ablation-eta-beta} for details). 

Therefore, EMA can serve as an \emph{automatic curriculum} \obs: it sets an initial small goal of $\frac{\tau}{1+\sigma^2}$ for $p_j$ so $\Delta_j$ need only be small and positive to both drive $p_j$ larger and satisfy Eqn.~\ref{eq:second-condition}. Then EMA gradually sets a higher goal for $p_j$ by increasing $\tau$, so that $p_j$ and $s_j$ can grow, while keeping the eigenspaces of $W_\pred$ and $F$ aligned. 

As a trade-off, a very slow EMA schedule ($\beta$ small) yields a slow training procedure \obs{} (See Fig.~\ref{fig:ablation-eta-beta}). Also small $\tau$ leads to larger $p^*_{j-}$ and more eigen modes can be trapped in the collapsed basin \obs. 

\vspace{-0.05in}
\subsection{Summarizing the effects of hyperparameters}
\vspace{-0.05in}
We summarize the positive and negative effects of multiple hyperparameters in Tbl.~\ref{tab:summarization-parameters}. We next provide additional ablations and experiments to further justify our reasoning. 

\textbf{Different weight decay $\eta_p$ and $\eta_s$}. If we set a higher weight decay for the predictor ($\eta_p$) than the online net ($\eta_s$), then $p_j$ grows slower than $s_j$ and it is possible that the condition of Theorem~\ref{thm:alignment-eigenspace} can still be satisfied without using EMA. Indeed Tbl.~\ref{tab:symmetric-no-EMA-works} shows this is the case. 

\begin{table}[]
    \centering
    \vspace{-0.1in}
    \small
    \setlength{\tabcolsep}{1.2pt}
    \begin{tabular}{c||c|c||c|c}
                 & \multicolumn{2}{c||}{No predictor bias} & \multicolumn{2}{c}{With predictor bias} \\
                 &  sym $W_\pred$ & regular $W_\pred$ & sym $W_\pred$ & regular $W_\pred$ \\
     \hline
     \multicolumn{5}{c}{\emph{Weight decay only for predictor ($\bar\eta_p = 0.0004$ and $\bar\eta_s = 0$)}} \\
     \hline
     EMA    &   $71.91{\pm}0.70$ & $70.54 {\pm}0.93$ & $73.67{\pm}0.47$  & $70.89{\pm}0.98$ \\
     no EMA &  $71.12 {\pm}0.71$ &  $71.34 {\pm}0.63$ & $73.01{\pm}0.37$ & $71.70{\pm}0.83$ \\
     \hline
     \multicolumn{5}{c}{\emph{No weight decay for all ($\bar\eta_p = \bar\eta_s = 0$)}} \\
     \hline
     EMA& $71.76{\pm}0.28$& $70.62{\pm}1.05$& $71.86{\pm}0.39$& $70.99{\pm}1.01$\\
     no EMA& {\color{red} $43.04{\pm}2.32$} & $71.36{\pm}0.44$& {\color{red} $41.36{\pm}3.33$} & $71.37{\pm}0.77$
    \end{tabular}
    \vspace{-0.05in}
    \caption{\small Symmetric weight works without EMA, if we set weight decay for the predictor ($\bar\eta_p=0.0004$) but not the trunk ($\bar\eta_s=0$) in BYOL experiment on STL-10. Report Top-1 accuracy after 100 epochs. If there is no weight decay for \emph{all layers}, then again symmetric weight doesn't work without EMA.}
    \label{tab:symmetric-no-EMA-works}
    %\vspace{-0.2in}
\end{table}

\begin{figure}
    \centering
    \includegraphics[width=0.45\textwidth]{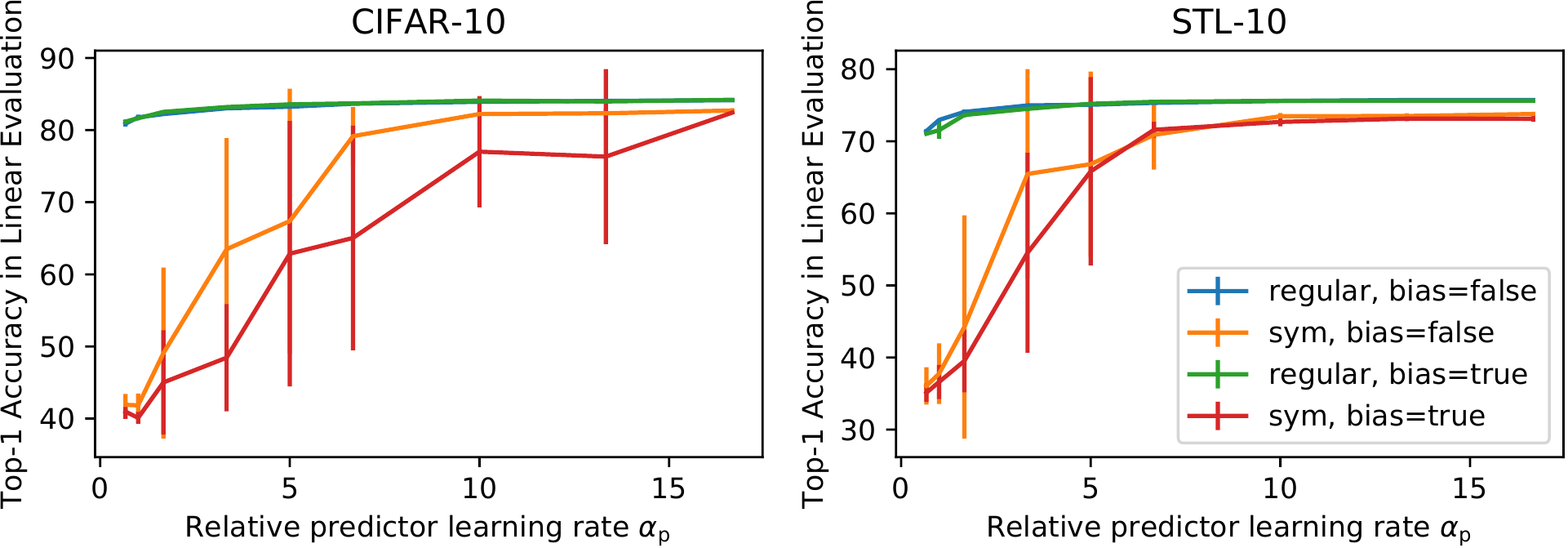}
    \vspace{-0.15in}
    \caption{\small The effects of relative learning rate $\alpha_\pred$ without EMA. If $\alpha_\pred > 1$, symmetric $W_\pred$ with no EMA can also work. Experiments on STL-10 and CIFAR-10~\cite{krizhevsky2009learning} (100 epochs with 5 random seeds).}
    \label{fig:alpha-pred}
\end{figure}

\textbf{Larger learning rate of the predictor $\alpha_\pred > 1$}. Our analysis predicts that one way to make symmetric $W_\pred$ work with no EMA is to use $\alpha_\pred > 1$ (i.e. Theorem~\ref{thm:alignment-eigenspace} is more easily satisfied). Fig.~\ref{fig:alpha-pred} verifies this prediction. Moreover Table 22 in Appendix of BYOL \cite{byol} also shows that $\alpha_\pred > 1$ is required to get BYOL working without EMA.  

As a reference, Table 22 in Appendix I.2 of BYOL~\cite{byol} also shows a similar trend: the learning rate of the (2-layer) predictor needs to be higher than that of the projector for strong performance in ImageNet, when EMA is absent.  

\vspace{-0.05in}
\section{Optimization-free Predictor $W_\pred$}
\label{sec:experiment-with-direct-pred}
\vspace{-0.05in}
A direct consequence of our theory is a new method for choosing the predictor that avoids gradient descent altogether. Instead, we estimate the correlation matrix $F$ of predictor inputs and directly set $W_\pred$ to be a function of this, thereby avoiding both the need to align the eigenspaces of $F$ and $W_\pred$ through optimization, and the need to initialize $W_\pred$ outside the basin of collapse. As we shall see, this exceedingly simple, theory motivated method also yields better performance in practice compared to gradient-based optimization of a linear predictor. 

We call our method \ourmethod{} which simply estimates $F$, computes its eigen-decomposition $\hat F = \hat U \hat \Lambda_F \hat U^\t$, where $\hat \Lambda_F = \diag[s_1, s_2, \ldots, s_d]$, and sets $W_p$ via
\begin{equation}
    p_j = \sqrt{s_j} + \epsilon \max_j s_j, \ \ W_\pred = \hat U\diag[p_j]\hat U^\t. \label{eq:direct-set-wp}
\end{equation}
This choice is theoretically motivated by eigenspace-alignment between $W_\pred$ and $F$ (Theorem.~\ref{thm:alignment-eigenspace}) and convergence to the invariant parabola $s_j \propto p_j^2$ in Eqn.~\ref{eq:s-p-relation} with weight decay ($\eta >0$).  
Here the estimate correlation matrix $\hat F$ can be obtained by a moving average:
\begin{equation}
   \hat F = \rho \hat F + (1 - \rho) \eee{B}{\vf\vf^\t} \label{eq:rho}
\end{equation}
where $\eee{B}{\cdot}$ is the expectation over a batch. Note that where $\vf$ is not zero-mean, we keep $\hat F$ a correlation matrix (rather than a covariance) \emph{without} zero-centering $\vf$, otherwise the performance deteriorates. We also added a regularization factor proportional to a small $\epsilon$ to boost the small eigenvalues $s_j$ so they can learn faster. In all our experiments on real-world datasets, we use $\ell_2$-normalization so the absolute magnitude of $s_j$ doesn't matter.

\textbf{Hyper-parameter \texttt{freq}}. Besides, we also evaluate a hybrid approach by introducing \texttt{freq}, which is how frequently eigen-decomposition is conducted for matrix $\hat F$ to set $W_p$. For example, \texttt{freq = 5} means that eigen decomposition is run every 5 minibatches. When $W_p$ is not set by eigen decomposition, it is updated by regular gradient updates. \texttt{freq = 1} means the eigen-decomposition is performed at every minibatch.  

\begin{table}[]
    \centering
    \small
    \setlength{\tabcolsep}{1pt}
    \begin{tabular}{c||c|c|c|c}
    & \multicolumn{4}{c}{Regularization factor $\epsilon$}\\
   & 0 & 0.01 & 0.1 & 0.5 \\
\hline
$\rho=0.3$ & $76.77{\pm}0.24$& $77.11{\pm}0.35$& $\mathbf{77.86{\pm}0.16}$ & $75.06{\pm}1.10$\\
$\rho=0.5$ & $76.65{\pm}0.20$& $76.76{\pm}0.33$& $\mathbf{77.56{\pm}0.25}$ & $75.22{\pm}0.81$
    \end{tabular}
    \vspace{-0.05in}
    \caption{\small STL-10 Top-1 after BYOL training for 100 epochs, if we use \ourmethod{} (Eqn.~\ref{eq:direct-set-wp}). It outperforms training $W_\pred$ using gradient descent ($74.51\%$ in Tbl.~\ref{tab:symmetric-no-EMA-does-not-work}, regular $W_\pred$ with EMA). EMA is used in all experiments. No predictor bias. $\rho$ defined in Eqn.~\ref{eq:rho}.}
    %\vspace{-0.2in}
    \label{tab:direct-set-wp}
\end{table}

\begin{table}[]
    \centering
    \small
    \setlength{\tabcolsep}{1pt}
    \begin{tabular}{c||c|c|c|c}
    & \multicolumn{4}{c}{Initial constant $c_j$}\\
& $0.1$ & $0.05$ & $-0.05$ & $-0.1$ \\
\hline
freq=1& $46.57{\pm}18.43$& $65.31{\pm}18.22$& $77.11{\pm}0.66$& $76.46{\pm}0.55$ \\ 
freq=2& $75.01{\pm}0.48$& $75.10{\pm}0.35$& $76.83{\pm}0.52$& $76.31{\pm}0.27$ 
    \end{tabular}
    \caption{\small STL-10 Top-1 Accuracy after BYOL training for 100 epochs. With different $c_j$. $\rho = 0.3$ and $\epsilon = 0$. EMA is used in all experiments. No predictor bias.}
    \vspace{-0.15in}
    \label{tab:different-cj}
\end{table}

Tbl.~\ref{tab:direct-set-wp} shows that directly computing $W_\pred$ through \ourmethod{} works \emph{better} ($76.77\%$) than training via gradient descent ($74.51\%$ in Tbl.~\ref{tab:symmetric-no-EMA-does-not-work}, regular $W_\pred$ with EMA). Additional regularization through $\epsilon$ yields even better performance ($77.38\%$). Different ways to estimate $F$ (moving average or simple average) yield only small differences.  

The performance of \ourmethod{} also remains good over many more training epochs (Tbl.~\ref{tab:accuracy-long-term}). Moreover, if we allow some gradient steps in between directly setting $W_\pred$ (i.e., \texttt{freq > 1}), performance becomes even better ($80.28\%$). This might occur because the estimated $\hat F$ may not be accurate enough and SGD can help correct it. This also mitigates the computational cost of eigen-decomposition. 

\textbf{The constant $c_j$}. What happens if $p_j = \sqrt{\max(s_j-c_j, 0)}$ with $c_j \neq 0$? If $c_j$ is small negative, performance is still fine but a positive $c_j$ leads to very poor performance (Tbl.~\ref{tab:different-cj}), likely due to many small eigen-values $s_j$ becoming zero and therefore trapped in the collapsed basin.

\textbf{Feature-dependent $W_\pred$}. Note one of the advantages of using two layer predictors is that $W_\pred$ can depend on the input features. We explored this idea by using a few random partitions of the input space, and within each random partition we estimated a different correlation matrix $\hat F$. The final $\hat F$ is the sum of all the correlation matrices.  With $6$ random partitions, \ourmethod{} achieves $78.20{\pm}0.16$ Top-1 accuracy after 100 epochs, closing performance gap to two-layer predictors ($78.85\%$ in Tbl.~\ref{tab:symmetric-no-EMA-does-not-work}). We leave a thorough analysis of the two layer setting to future work.

\begin{table}[t]
    \centering
    \small
    \setlength{\tabcolsep}{1pt}
    \begin{tabular}{c||c|c|c}
    & \multicolumn{3}{c}{Number of epochs}\\
                        &   100 & 300 & 500 \\
     \hline 
     \multicolumn{4}{c}{\emph{STL-10}} \\
     \hline
     \ourmethod{}  & $\mathbf{77.86{\pm}0.16}$  & $78.77{\pm}0.97$  & $78.86 {\pm}1.15$ \\ 
     \ourmethod{} (freq=5) & $77.54 {\pm}0.11$  & $\mathbf{79.90{\pm}0.66}$  & $\mathbf{80.28{\pm}0.62}$ \\
     SGD baseline  &    $75.06{\pm}0.52$  & $75.25 {\pm}0.74$  &  $75.25 {\pm}0.74$ \\
     \hline 
     \multicolumn{4}{c}{\emph{CIFAR-10}} \\
     \hline
    \ourmethod{}          & $\mathbf{85.21{\pm}0.23}$  & $\mathbf{88.88{\pm}0.15}$ & $89.52{\pm}0.04$ \\
    \ourmethod{} (freq=5) & $84.93{\pm}0.29$ & $88.83{\pm}0.10$ & $\mathbf{89.56{\pm}0.13}$ \\
    SGD baseline          & $84.49{\pm}0.20$ & $88.57{\pm}0.15$ & $89.33{\pm}0.27$
     \end{tabular}
    \vspace{-0.1in}
    \caption{\small STL-10/CIFAR-10 Top-1 accuracy of \ourmethod{}, after training for longer epochs. $\rho=0.3$, $\epsilon=0.1$ with EMA. }
    \label{tab:accuracy-long-term}
\end{table}

\textbf{ImageNet experiments}. We conducted additional experiments on ImageNet~\cite{imagenet_cvpr09}, with our own BYOL~\cite{byol} implementation. We used ResNet-50~\cite{resnet} as the backbone to produce features for a linear probe, followed by a projector and a predictor. The architecture design (e.g., feature dimensions), augmentation strategies (e.g., color jittering, blur~\cite{simclr}, solarization, etc.) and linear classification protocol strictly follow BYOL~\cite{byol}. 

We experimented with two different training settings to study the generalization ability of \ourmethod{}. In the first setting, we employ an asymmetric loss (given two views, only one view is used as the prediction target). The loss is optimized using standard SGD for 60 epochs with a batch size of 256. The second setting follows BYOL more closely, where we use a symmetrized loss, 4096 batch size and LARS optimizer~\cite{you2017large}, and train for 300 epochs.

The results are summarized in Tbl.~\ref{tab:accuracy-on-imagenet}. Both settings exhibit similar behaviors in comparison, and we take the 300-epoch results as our highlights in the following. As a baseline, the default 2-layer predictor from BYOL (with BatchNorm and ReLU, 4096 hidden dimension, 256 input/output dimension) achieves 72.5\% top-1 accuracy, and 90.8\% top-5 accuracy with 300-epoch pre-training. This reproduces the accuracy reported in BYOL~\cite{byol}. We find \ourmethod{} can match this performance (72.4\% top-1, and 91.0\% top-5) \emph{without} any gradient-based training by instead directly setting the (256$\times$256) linear predictor weights every mini-batch. In particular for top-5 \ourmethod{} is even 0.2\% better. For a fair comparison, we also run BYOL with a learned linear predictor. We find the performance drops to 69.9\%, and 89.6\% respectively (2.5\% gap to our method). The gap is even bigger in 60-epoch settings, up to 5.0\% in top-1 (59.4\% vs. 64.4\%). These experiments demonstrate the success of \ourmethod{} on STL-10 and CIFAR can also generalize and scale to ImageNet.

\section{Discussion}

\paragraph{Summary.} Therefore, remarkably, our theoretical analysis of non-contrastive SSL, primarily centered around a $3$ dimensional nonlinear dynamical system, not only yields conceptual insights into the functional roles of complex ingredients like EMA, stop-gradients, predictors, predictor symmetry, diverse learning rates, weight decay and all their interactions, but also predicts the performance patterns of many ablation studies as well as suggests an exceedingly simple \ourmethod{} method that rivals the performance of more complex predictor dynamics in real-world settings.

\textbf{Two-layer non-linear predictor}. 
With only a linear predictor, our results on ImageNet (Tbl.~\ref{tab:accuracy-on-imagenet}) have already shown strong performance, on par with a default BYOL setting with a 2-layer predictor on ImageNet. One interesting question is how the dynamics changes if the predictor has 2 layers. While we don't provide a formal analysis and the math can be quite complicated, the intuition here is that the ``fat'' 2-layer predictor used in practice (e.g., more (4096) hidden dimension than input/output dimensions (256), and a ReLU in between) essentially provides a large pool of initial weight directions to start with, and some of them could be  ``lucky draws'', that make eigen-space alignment faster. On the other hand, a 1-layer predictor with gradient updates may get stuck in local minima. Therefore, with the same number of epochs, a 2-layer predictor outperforms 1-layer, and is comparable with \ourmethod{} which does not suffer from local minima issues.

\section*{Acknowledgements} We thank Lantao Yu for helpful discussions.

\begin{table}[t]
    \centering
    \small
    \setlength{\tabcolsep}{6pt}
    \begin{threeparttable}
    \begin{tabular}{l||c|c||c|c}
     \multirow{2}{*}{BYOL variants} & \multicolumn{2}{c||}{\emph{Accuracy (60 ep)}} & \multicolumn{2}{c}{\emph{Accuracy (300 ep)}} \\
            & Top-1 & Top-5 & Top-1 & Top-5 \\
     \hline  
     2-layer predictor\tnote{*} &   $\mathbf{64.7}$  &  $\mathbf{85.8}$ & $\mathbf{72.5}$  & $90.8$ \\
     linear predictor  &    $59.4$ & $82.3$ & $69.9$  & $89.6$ \\
     \ourmethod{}  & $64.4$  & $\mathbf{85.8}$ & $72.4$ & $\mathbf{91.0}$ \\
     \end{tabular}
     \begin{tablenotes}
        \footnotesize
        \item[*] 2-layer predictor is BYOL default setting.
     \end{tablenotes}
    \end{threeparttable}
    \caption{\small ImageNet experiments comparing \ourmethod{} with BYOL~\cite{byol}. \emph{Without} gradient-based training, \ourmethod{} is able to match the performance of the default 2-layer predictor introduced by BYOL, and significantly outperform the linear predictor by $5\%$ (60 epoch) and $2.5\%$ (300 epoch).}
    \label{tab:accuracy-on-imagenet}
\end{table}

\newpage

\bibliographystyle{icml2021}
\bibliography{references}

\onecolumn
\appendix 

\setcounter{lemma}{0}
\setcounter{theorem}{0}

\def\stopgrad{\mathrm{StopGrad}}

\icmltitle{Supplementary Materials for ``\ourtitle{}''}

\section{Section~\ref{sec:two-layer-model}}
\begin{lemma}[Dynamics of BYOL/SimSiam]
For objective ($\vf_1 = W\vx_1$ and $\vf_{2\ema} = W_\ema \vx_2$ where $W_\ema$ is EMA weight): 
\begin{equation}
    J(W, W_\pred) := \frac{1}{2}\eee{\vx\sim p(\cdot),\ \ \vx_1, \vx_2\sim p_\aug(\cdot|\vx)}{\|W_\pred \vf_1 - \stopgrad(\vf_{2\ema})\|^2_2}
\end{equation}

Let $X = \ee{\bar\vx\bar\vx^\t}$ where $\bar\vx(\vx) := \eee{\vx'\sim p_\aug(\cdot|\vx)}{\vx'}$ is the average augmented view of a data point $\vx$ and  $X' = \eee{\vx}{\var_{\vx'|\vx}[\vx']}$ is the covariance matrix $\var_{\vx'|\vx}[\vx']$ of augmented views $\vx'$ conditioned on $\vx$, subsequently averaged over the data $\vx$. The dynamics is the following:
\begin{eqnarray}
    \dot W_p &=& -\frac{\partial J}{\partial W_p} = - W_p W(X+X')W^\t + W_\ema X W^\t \\
    \dot W &=& - \frac{\partial J}{\partial W} = - W_p^\t W_p W (X+X') + W_p^\t W_\ema X
\end{eqnarray}
\end{lemma}
\begin{proof}
Note that 
\begin{eqnarray}
    & & (W_\pred \vf_1 - \vf_{2\ema})^\t (W_\pred \vf_1 - \vf_{2\ema}) \\
    &=& \vf^\t_1 W_\pred^\t W_\pred \vf_1 - \vf^\t_{2\ema} W_\pred \vf_1 - \vf_1^\t W_\pred^\t \vf_{2\ema} + \vf_{2\ema}^\t \vf_{2\ema} \\
    &=& tr(W_p^\t W_p \vf_1\vf_1^\t) - tr(W_p \vf_1\vf_{2\ema}^\t) - tr(W_p^\t \vf_{2\ema}\vf_1^\t) + tr(\vf_{2\ema}\vf_{2\ema}^\t)
\end{eqnarray}
Let $F_1 = \ee{\vf_1\vf_1^\t} = W(X+X')W^\t$ where $X = \eee{\vx}{\bar\vx\bar\vx^\t}$ and $X'=\eee{\vx}{\var_{\vx'|\vx}[\vx']}$, $F_{1,2\ema} = \ee{\vf_1\vf_{2\ema}^\t}$, $F_{2\ema,1} = \ee{\vf_{2\ema}\vf_1^\t} = F_{1,2\ema}^\t$ and $F_{2\ema} = \ee{\vf_{2\ema}\vf^\t_{2\ema}}$. This leads to:
\begin{eqnarray}
    J(W, W_\pred) &=& \frac{1}{2}\left[tr(W_p^\t W_p F_1) - tr(W_p F_{1,2\ema}) - tr(F_{1,2\ema}W_p) + tr(F_{2\ema})\right]
\end{eqnarray}

Taking partial derivative with respect to $W_p$ and we get the gradient update rule:
\begin{equation}
    \dot W_p = -\frac{\partial J}{\partial W_p} = - W_p F_1 + F_{1,2\ema}^\t
\end{equation}
Now we take the derivative with respect to $W$. Note that we have stop-gradient in $\vf_{2\ema}$, so we would like to be careful when taking derivatives. We first compute $\partial J / \partial F_1$ and $\partial J / \partial F_{1,2\ema}$. Note that both $F_1$ and $F_{1,2\ema}$ contains $W$, due to the fact that we have stop gradient, $F_1$ is a quadratic form of $W$ but $F_{1,2\ema}$ is a \emph{linear} form of $W$. This is critical.  
\begin{eqnarray}
    \frac{\partial J}{\partial F_1} &=& \frac{1}{2}W_p^\t W_p \\ 
    \frac{\partial J}{\partial F_{1,2\ema}} &=& -W^\t_p
\end{eqnarray}
Let $W = [w_{ij}]$ and $X = \ee{\bar\vx\bar\vx^\t}$ ($X_\tot$ and $X'$ are defined similarly). We have $F_1 = W(X+X')W^\t$ and $F_{1,2\ema} = WXW^\t_\ema$. So we have:
\begin{equation}
\frac{\partial J}{\partial w_{ij}} = \sum_{kl} \left[\frac{\partial J}{\partial F_1}\right]_{kl} \frac{\partial [F_1]_{kl}}{\partial w_{ij}} + \sum_{kl} \left[\frac{\partial J}{\partial F_{1,2\ema}}\right]_{kl} \frac{\partial [F_{1,2\ema}]_{kl}}{\partial w_{ij}}
\end{equation}
Let $C = X + X'$, here we have:
\begin{eqnarray}
    & & \sum_{kl} \left[\frac{\partial J}{\partial F_1}\right]_{kl} \frac{\partial [F_1]_{kl}}{\partial w_{ij}} = \sum_{kl} \left[\frac{\partial J}{\partial F_1}\right]_{kl} \sum_{mn} \frac{\partial w_{km}c_{mn}w_{ln}}{\partial w_{ij}} \\
    &=& \sum_{kl} \left[\frac{\partial J}{\partial F_1}\right]_{kl} \left(\delta(i=k)\sum_n c_{jn}w_{ln} + \delta(i=l)\sum_m w_{km}c_{mj}\right) \\
    &=& \sum_l \left[\frac{\partial J}{\partial F_1}\right]_{il} \sum_n c_{jn}w_{ln} + \sum_k \left[\frac{\partial J}{\partial F_1}\right]_{ki} \sum_m w_{km}c_{mj} \\
    &=& \left[\frac{\partial J}{\partial F_1}WC^\t + \frac{\partial J}{\partial F^\t_1} WC\right]_{ij}
\end{eqnarray}
Similarly (note that we don't take derivative with respect to $W_\ema$):
\begin{eqnarray}
    \sum_{kl} \left[\frac{\partial J}{\partial F_{1,2\ema}}\right]_{kl} \frac{\partial [F_{1,2\ema}]_{kl}}{\partial w_{ij}} = \left[\frac{\partial J}{\partial F_{1,2\ema}}W_\ema X^\t\right]_{ij}
\end{eqnarray}

So we have:
\begin{equation}
    \dot W = - \frac{\partial J}{\partial W} = - W_p^\t W_p W (X+X') + W_p^\t W_\ema X
\end{equation}

After some manipulation, we finally arrive at the following gradient update rule:
\begin{eqnarray}
    \dot W_p &=& [- W_p W(X+X') + W_\ema X] W^\t \label{eq:grad-wp-appendx-no-decay} \\
    \dot W &=& W^\t_p [- W_p W (X+X') + W_\ema X] \label{eq:grad-w-appendix-no-decay}
\end{eqnarray}
\end{proof}
\textbf{Remarks}. For symmetric loss:
\begin{equation}
    J(W, W_\pred) := \frac{1}{4}\eee{\vx\sim p(\cdot),\ \ \vx_1, \vx_2\sim p_\aug(\cdot|\vx)}{\|W_\pred \vf_1 - \stopgrad(\vf_{2\ema})\|^2_2 + \|W_\pred \vf_2 - \stopgrad(\vf_{1\ema})\|^2_2}
\end{equation}
The update rule is done by swapping subscript $1$ and $2$ in the update rule of $W_p$ (here $F_2 = \ee{\vf_2\vf_2^\t}$): 
\begin{equation}
    \dot W_p = -\frac{\partial J}{\partial W_p} = - \frac12 W_p(F_1 + F_2) + \frac{1}{2}(F_{2\ema,1} + F_{1\ema,2})
\end{equation}
Under the large batch limit, it is the same as Eqn.~\ref{eq:grad-wp-appendx}.

Note that the Lemma doesn't include weight decay. With weight decay $\eta$, it is not hard to see that we will arrive at the following slightly altered gradient flow:
\begin{eqnarray}
    \dot W_p &=& [- W_p W(X+X') + W_\ema X] W^\t - \eta W_p \label{eq:grad-wp-appendx} \\
    \dot W &=& W^\t_p [- W_p W (X+X') + W_\ema X] - \eta W \label{eq:grad-w-appendix}
\end{eqnarray}

\begin{theorem}[Invariance of the Gradient Update]
The gradient update rules (Eqn.~\ref{eq:grad-wp} and Eqn.~\ref{eq:grad-w}) has the following invariance (where the symmetric matrix $C$ depends on initialization):
\begin{equation}
    W(t)W^\t(t) =  W_p^\t(t) W_p(t) + e^{-2\eta t}C 
\end{equation}
\end{theorem}
\begin{proof}
From Eqn.~\ref{eq:grad-w-appendix} and Eqn.~\ref{eq:grad-wp-appendx}, we know that
\begin{equation}
   \alpha_\pred^{-1} W_p^\t \dot W_p + \alpha_\pred^{-1}\eta W_p^\t W_p = \dot W W^\t + \eta WW^\t 
\end{equation}
Taking transpose and we have:
\begin{equation}
    \alpha_\pred^{-1}\dot W^\t_p W_p + \alpha_\pred^{-1}\eta W_p^\t W_p = W \dot W^\t + \eta WW^\t 
\end{equation}
Adding them together and multiply both side with $e^{2\eta t}$:
\begin{equation}
    \alpha_\pred^{-1}\frac{\dd}{\dd t} (e^{2\eta t} W_p^\t W_p) = \frac{\dd}{\dd t}(e^{2\eta t} WW^\t) 
\end{equation}
This leads to $e^{2\eta t} WW^\t = \alpha_\pred^{-1} e^{2\eta t} W_p^\t W_p +C$, or $WW^\t = \alpha_\pred^{-1}W_p^\t W_p + e^{-2\eta t}C$.
\end{proof}

\begin{lemma}[Dynamics of a negative definite system]
\label{lemma:nd-decay}
Let $H(t)$ be $d$-by-$d$ time-varying positive definite (PD) matrices whose minimal eigenvalues are bounded away from 0: $\inf_{t\ge 0} \lambda_{\min}(H(t)) \ge \lambda_0 > 0$, then the following dynamics:
\begin{equation}
\frac{\dd \vw(t)}{\dd t} = -H(t)\vw(t)
\end{equation}
satisfies $\|\vw(t)\|_2 \le e^{-\lambda_0 t} \|\vw(0)\|_2$, which means that $\vw(t) \rightarrow 0$.
\end{lemma}
\begin{proof}
Construct the following Lyapunov function $V(\vw) := \frac{1}{2}\|\vw\|^2_2$. For $V(\vw(t))$ we have:
\begin{equation}
    \frac{\dd V}{\dd t} = \frac{\dd V}{\dd \vw} \frac{\dd \vw}{\dd t} =  -\vw^\t(t) H(t)\vw(t) 
\end{equation}
Note that $H(t)$ has eigen-decomposition: $H(t) = \sum_j \lambda_j(t)\vu_j(t)\vu^\t_j(t)$ with all $\lambda_j(t) \ge \lambda_0$ and $[\vu_1(t), \vu_2(t), \ldots, \vu_d(t)]$ forming an orthonormal bases. Therefore:
\begin{equation}
    \vw^\t H\vw = \sum_j \lambda_j \vw^\t\vu_j\vu^\t_j\vw\ge \lambda_0 \vw^\t\left[\sum_j \vu_j\vu^\t_j\right]\vw = \lambda_0 \|\vw\|^2_2
\end{equation}
Therefore, we have:
\begin{equation}
    \frac{\dd V}{\dd t} \le -\lambda_0 \|\vw(t)\|_2^2 = -2\lambda_0 V 
\end{equation}
which leads to $V(t) \le e^{-2\lambda_0 t}V(0)$. That is $\|\vw(t)\|_2 \le e^{-\lambda_0 t} \|\vw(0)\|_2$.
\end{proof}

\begin{theorem}[No-stop gradient will not work]
With $W_\ema=W$ (SimSiam case), removing the stop-gradient signal yields a gradient update for $W$ given by positive semi-definite (PSD) matrix $H(t) := X'\otimes(W_p^\t W_p+I) + X\otimes\tilde W_p^\t \tilde W_p$ (here $\tilde W_p := W_p - I$ and $\otimes$ is the Kronecker product):  
\begin{equation}
    \frac{\dd}{\dd t}\vec(W) = - H(t) \vec(W).  
\end{equation}
If $\inf_{t\ge 0} \lambda_{\min}(H(t)) \ge \lambda_0 > 0$, then $W(t)\rightarrow 0$. 
\end{theorem}
\begin{proof}
Note that if we don't have stop gradient and $W_\ema = W$, then we have additional terms (and we also need to compute $\partial J / \partial F_2$). Let $\tilde W_p = W_p - I_{n_2}$ and we have:
\begin{eqnarray}
    \dot W &=& - \frac{\partial J}{\partial W} = - W_p^\t W_p W (X+X') + (W_p^\t + W_p) WX - W(X+X') - \eta W \\
    &=& - (W_p^\t W_p + I) WX' - (W_p^\t W_p - W_p^\t - W_p + I) WX - \eta W \\
    &=& - (W_p^\t W_p + I) WX' - (W_p - I)^\t (W_p - I) WX - \eta W \\
    &=& - (W_p^\t W_p + I) WX' - \tilde W_p^\t \tilde W_p WX - \eta W \label{eq:full-grad}
\end{eqnarray}
With $\vec(AXB) = (B^\t \otimes A)\vec(X)$ and we see:
\begin{equation}
    \frac{\dd}{\dd t}\vec(W) = - \left[X' \otimes (W_p^\t W_p + I) + X\otimes \tilde W_p^\t \tilde W_p + \eta I_{n_1n_2} \right] \vec(W)  \label{eq:full-grad}
\end{equation}

If $\inf_{t\ge 0} \lambda_{\min}(H(t)) \ge \lambda_0 > 0$, then applying Lemma~\ref{lemma:nd-decay} and we have $\|\vec(W(t))\|_2 \le e^{-\lambda_0 t}\|\vec(W(0))\|_2 \rightarrow 0$, and there is no chance for $W$ to learn any meaningful features. 
\end{proof}
\textbf{Remark.} Note that if $W_a = W$ and we choose not to use the predictor ($W_p = I$), then no matter whether we choose to use stop-gradient or not, $W(t)$ always goes to $0$. The theorem above already proved that without stop gradient, it is the case. When there is stop gradient, from Eqn.~\ref{eq:grad-w}, we have:
\begin{equation}
    \dot W = - (X' + \eta I) W 
\end{equation}
Note that $X' + \eta I$ is a PD matrix and with similar arguments, $W(t)\rightarrow 0$.

\section{Section~\ref{sec:assumption}}
\textbf{Isometric assumptions.} Now we use the assumption that $X = I$ and $X'=\sigma^2 I$, which leads to 
\begin{equation}
    \dot F = \dot WXW^\t + WX\dot W^\t = -(1+\sigma^2)(W_p^\t W_p F + F W_p^\t W_p) + W_p^\t W_\ema W^\t + WW_\ema^\t W_p
\end{equation}
here $F = WXW^\t = WW^\t$. If we also have weight decay $-\eta W$ for $W$, then we have:
\begin{equation}
    \dot F = -(1+\sigma^2)(W_p^\t W_p F + F W_p^\t W_p) + W_p^\t W_\ema W^\t + WW_\ema^\t W_p - 2\eta F 
\end{equation}
or using anticommutator $\{A,B\} := AB+BA$:
\begin{equation}
    \dot F = -(1+\sigma^2)\{F, W_p^\t W_p\} + W_p^\t W_\ema W^\t + WW^\t_\ema W_p - 2\eta F 
\end{equation}
Similarly, for $W_p$ we have:
\begin{equation}
    \dot W_p = -\alpha_p (1+\sigma^2) W_p F + \alpha_p \tau F - \eta W_p
\end{equation}

\textbf{EMA assumption (Assumption~\ref{assumption:ema})}. Now we further study the effect of EMA. To model it, we just let $W_\ema = \tau W$ where $\tau < 1$ is a coefficient that measure how much EMA attenuates $W$. If $\tau = 1$ then $W_\ema = W$ and there is no EMA. Note that $\tau$ is not the same as the EMA parameter $1 - \gamma_\ema$, which is often set to be a fixed $0.004$ (or $1-0.996$). Instead, $\tau = \tau(t)$ is a changing parameter depends on how quickly $W=W(t)$ grows over time. If $W$ remains stable, then $\tau \approx 1$; if $W$ grows rapidly, then $\tau$ becomes small. 

\begin{figure}
    \centering
    \includegraphics[width=0.33\textwidth]{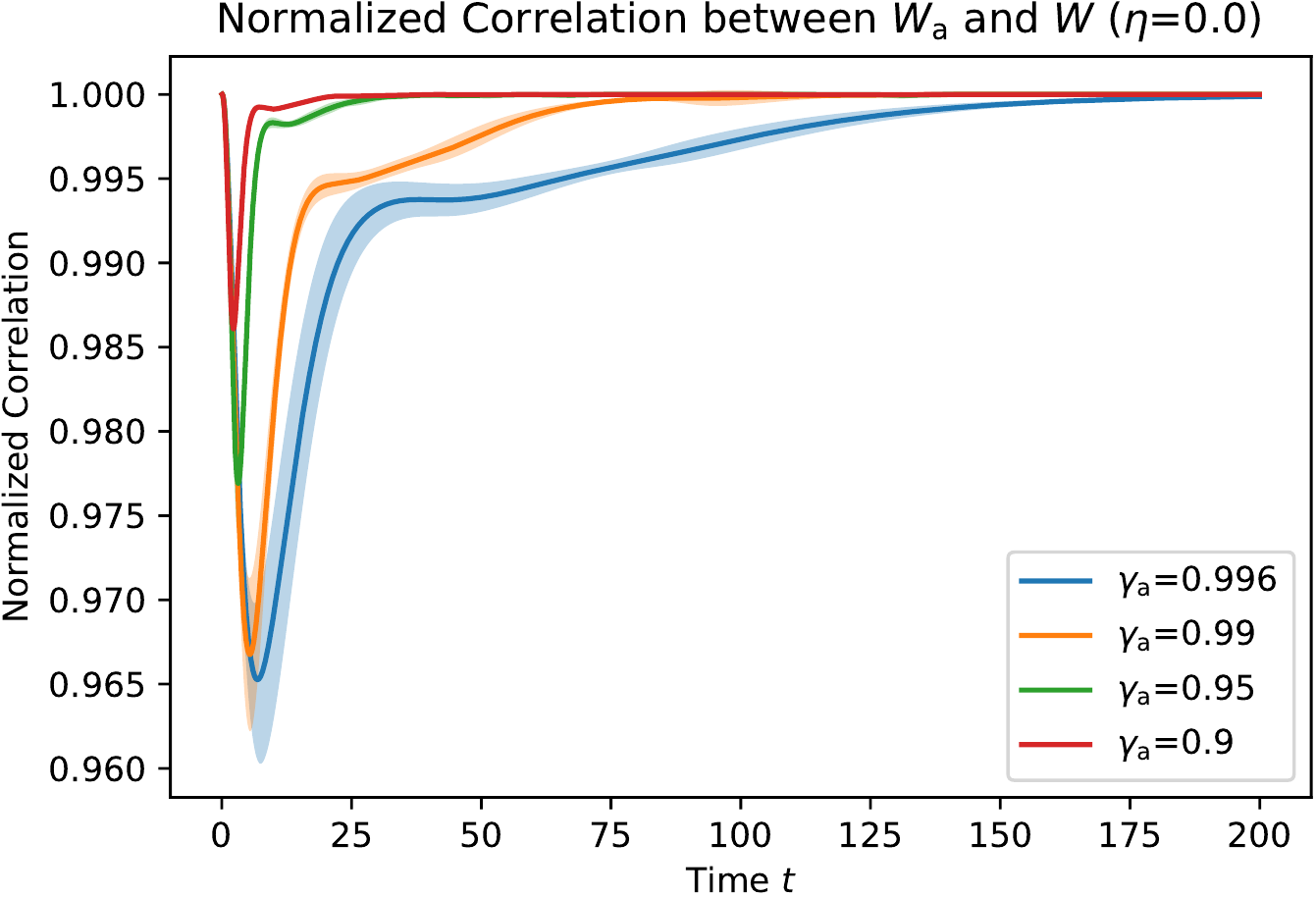}
    \includegraphics[width=0.33\textwidth]{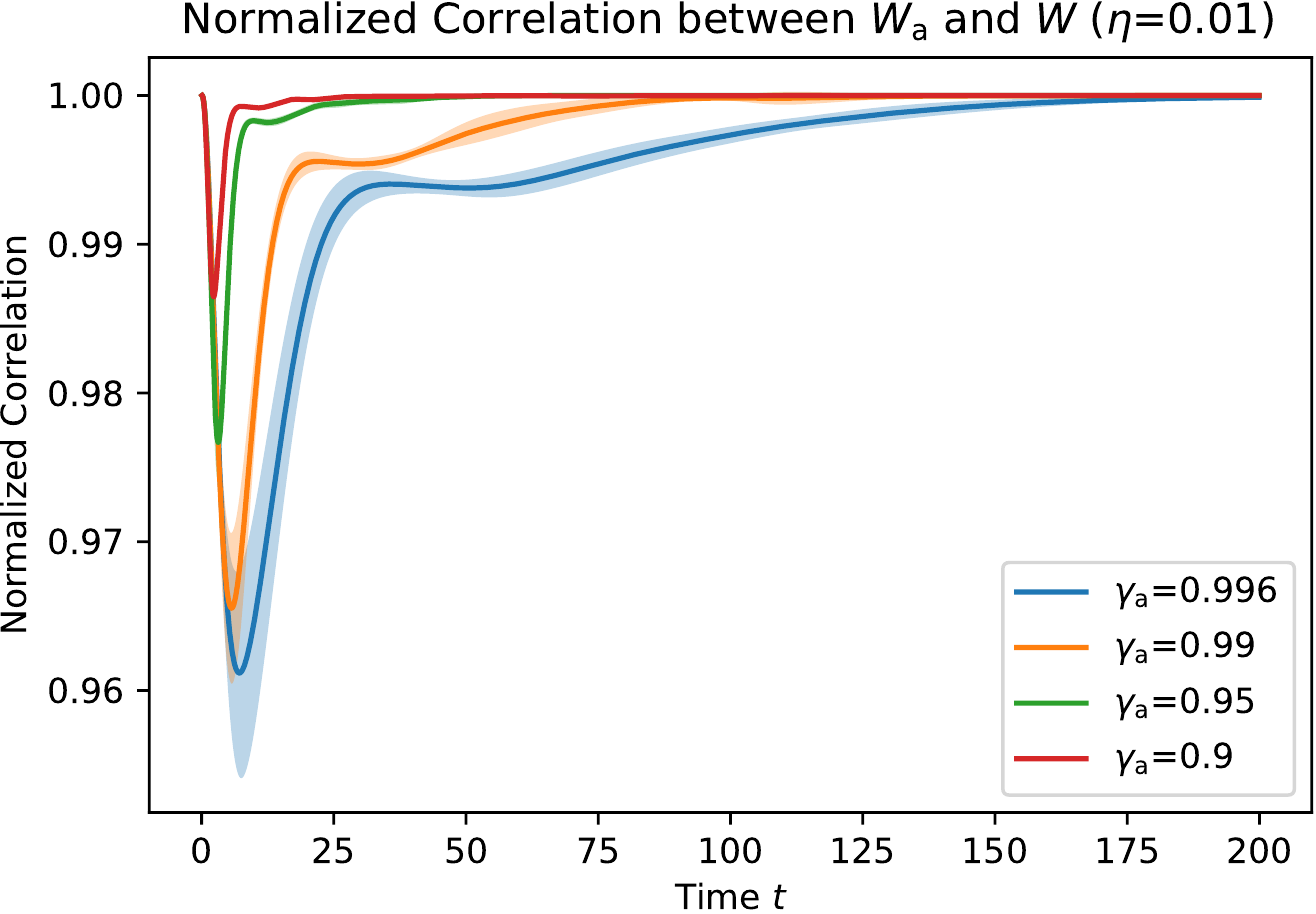}
    \includegraphics[width=0.33\textwidth]{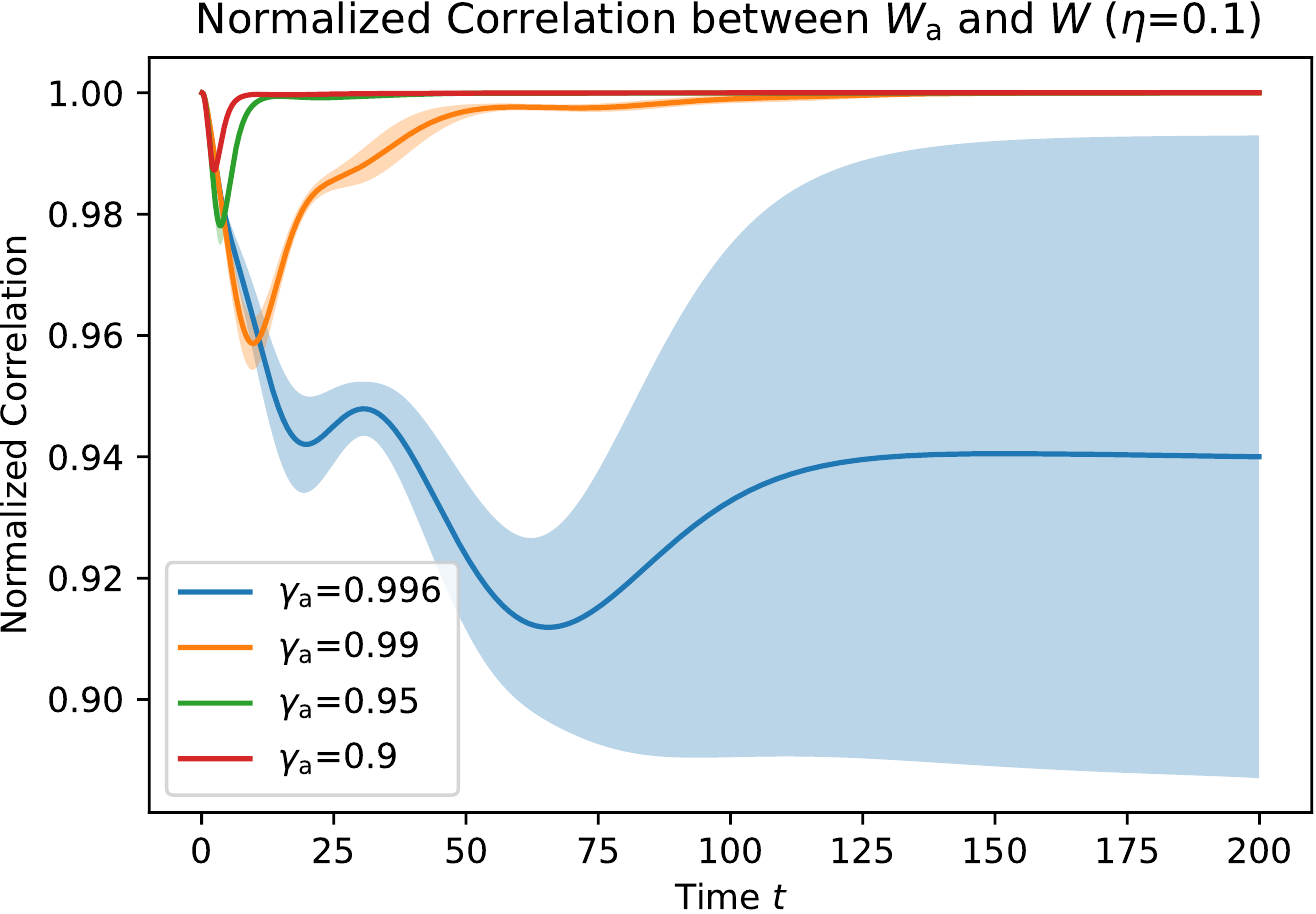}
    \caption{Check the validity of EMA assumption (Assumption~\ref{assumption:ema}) with different EMA coefficients $\gamma_\ema$ for BYOL dynamics with $X=I$ and $X'=\sigma^2 I$ (Assumption~\ref{assumption:isotropic}). $\sigma = 0.03$. All experiments are run $10$ times to get mean and standard derivation (shaded area). We could see the EMA assumption is largely correct. Even at the region with $\gamma_\ema$ close to $1$ (e.g., $0.996$) and large $\eta$, the normalized correlation between $W_\ema$ and $W$ are still high ($\sim 0.9$). Note that throughout our analysis, the initial value of $W_\ema(0) = 0$. \textbf{Left:} weight decay $\eta=0$, \textbf{Middle:} $\eta=0.01$, \textbf{Right:} $\eta=0.1$.}
    \label{fig:check-ema-assumption}
\end{figure}

Fig.~\ref{fig:check-ema-assumption} shows that this assumption is largely correct. 

Under this condition, using $F = WXW^\t = WW^\t$, the dynamics becomes (Now we also put weight decay for $W_p$):
\begin{eqnarray}
    \dot W_p &=& - \alpha_\pred(1+\sigma^2) W_p F + \alpha_\pred\tau F - \eta W_p \label{eq:wp-assumption-1-2}\\
    \dot F &=& -(1+\sigma^2)(W_p^\t W_p F + F W_p^\t W_p) + \tau  (W_p^\t F + F W_p) - 2\eta F 
\end{eqnarray}

\textbf{Derivation of Fixed point of Eqn.~\ref{eq:grad-wp}}. Given the dynamics Eqn.~\ref{eq:wp-assumption-1-2} we now want to check its fixed point:
\begin{equation}
    - \alpha_\pred(1+\sigma^2) W_p F + \alpha_\pred\tau F - \eta W_p = 0
\end{equation}
for some PSD matrix $F$. For convenience, let $\eta' = \eta / \alpha_\pred$. Since $F$ is always PSD, we have eigendecomposition $F = U\Lambda U^\t$. Left-multiplying $U$ and right-multiplying $U^\t$, we have:
\begin{equation}
    (1+\sigma^2) \bar W_p \Lambda + \eta' \bar W_p = \tau \Lambda 
\end{equation}
where $\bar W_\pred := U^\t W_\pred U$. Let $\Lambda' = (1+\sigma^2) \Lambda + \eta'I$ is a diagonal matrix with all positive diagonal element since $\eta' > 0$. Therefore, we have:
\begin{equation}
    \bar W_p \Lambda' = \tau \Lambda
\end{equation}
and thus $\bar W_p = \tau \Lambda (\Lambda')^{-1}$ is a symmetric matrix and so does $W_\pred = U\bar W_\pred U^\t$. When $\eta = 0$ and $F$ has zero eigenvalues, $W_\pred$ can have infinite solutions (or fixed points), and some of them might not be symmetric.

\textbf{Symmetrization of $W_p$}. Now we need to assume $W_p$ is symmetric and also symmetrize its dynamics, which yields (here $\{A,B\}:= AB+BA$):
\begin{eqnarray}
    \dot W_\pred &=& -\frac{\alpha_\pred}{2}(1+\sigma^2)\{W_\pred, F\} + \alpha_\pred \tau F - \eta W_\pred \label{eq:symmetric-appendix}\\
    \dot F &=& -(1+\sigma^2) \{W_\pred^2, F\}+ \tau\{W_\pred,F\} - 2\eta F \nonumber
\end{eqnarray}

Note that the asymmetric dynamic might be interesting and we will leave it later. 

\def\vlambda{\boldsymbol{\lambda}}
\def\vone{\boldsymbol{1}}

\subsection{Section~\ref{sec:eigen-alignment}}
\begin{theorem}[Alignment of Eigenspace]
Under the dynamics of Eqn.~\ref{eq:symmetric-appendix}, the commutator $[F,W_\pred] := F W_p - W_p F$ satisfies:
\begin{equation}
    \frac{\dd}{\dd t} [F,W_\pred] = -[F,W_\pred] K - K [F,W_\pred] 
\end{equation}
where 
\begin{equation}
K = K(t) = (1+\sigma^2)\left[\frac{\alpha_\pred}{2}F(t) + W_p^2(t) - \frac{\tau}{1+\sigma^2} W_p(t)\right] + \frac{3}{2}\eta I \label{eq:K-appendix}
\end{equation}
If $\max_{t\ge 0}\lambda_{\min}[K(t)] = \lambda_0 > 0$, then the commutator $\|[F(t),W_\pred(t)]\|_F \le e^{-2\lambda_0 t} \|[F(0),W_\pred(0)]\|_F \rightarrow 0$, i.e., the eigenspace of $W_\pred$ gradually aligns with $F$. 
\end{theorem}

\begin{figure}
    \centering
    \includegraphics[width=\textwidth]{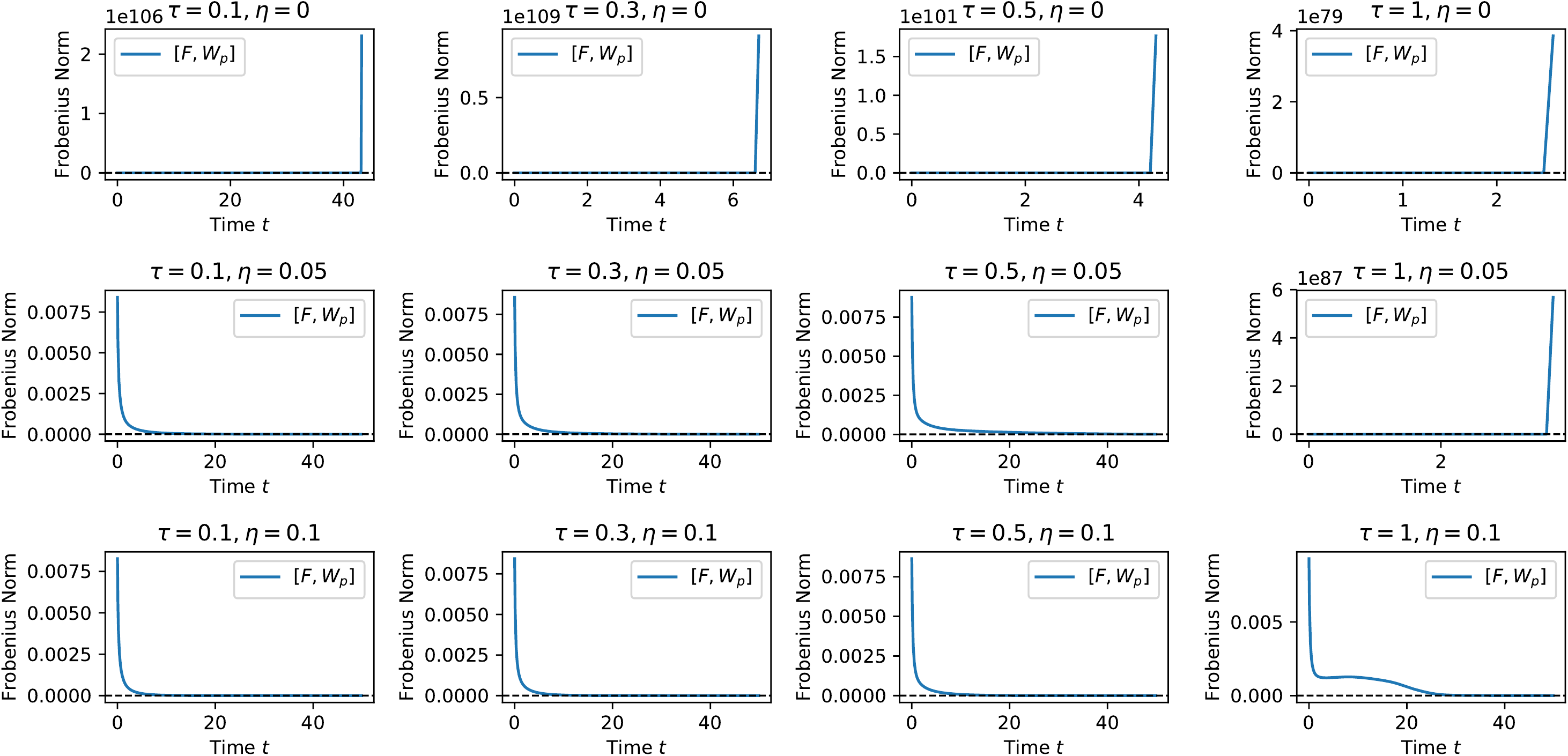}
    \caption{\small The norm of the communicator $[F,W_p]$ over time under different hyper-parameters (different \emph{time-independent} $\tau$ and different weight decay $\eta$) in symmetrized dynamics Eqn.~\ref{eq:symmetric-appendix}. When weight decay is small or zero, and/or $\tau$ is large, the norm of the communicator $\|[F,W_p]\|_F$ can shoot up (no eigenspace alignment). } 
    \label{fig:verify-eigen-alignment}
\end{figure}

\begin{proof}
Let's compute the commutator $L := [F,W_\pred] := F W_p - W_p F$ and its time derivative. First we have:
\begin{equation}
    F\dot W_p - \dot W_p F = - \frac{\alpha_\pred}{2}(1+\sigma^2)(FL + L F) - \eta L
\end{equation}
Then we have
\begin{equation}
    \dot F W_p - W_p \dot F = -(1+\sigma^2)(W_p^2 L + L W_p^2) + \tau  (W_p L + L W_p) - 2\eta L
\end{equation}
So we have
\begin{equation}
    \dot L = F\dot W_p + \dot F W_p - (W_p \dot F + \dot W_p F) = -KL - LK 
\end{equation}
where 
\begin{equation}
K = K(t) = (1+\sigma^2)\left[\frac{\alpha_\pred}{2}F + W_p^2 - \frac{\tau}{1+\sigma^2} W_p\right] + \frac32\eta I \label{eq:K-appendix}
\end{equation}
is a symmetric matrix. We can write the dynamics of $L(t)$: 
\begin{equation}
    \frac{\dd \vec(L(t))}{\dd t} = - \left[K(t) \oplus K(t)\right] \vec(L(t))
\end{equation}
where $K(t) \oplus K(t) := I\otimes K(t) + K(t) \otimes I$ is the Kronecker sum and is a PSD matrix if $K$ is PSD.  

If $\inf_{t\ge 0}\lambda_{\min}(K(t)) \ge \lambda_0 > 0$ for all $t$, then $\inf_{t\ge 0}\lambda_{\min}[K(t)\oplus K(t)] \ge 2\lambda_0$. Applying Lemma~\ref{lemma:nd-decay} and we have:
\begin{equation}
\|\vec(L)\|_2 \le  e^{-2\lambda_0 t} \|\vec(L(0))\|_2 \rightarrow 0    
\end{equation}
This means that $W_p$ and $F$ can commute, and the eigen space of $W_p$ and $F$ will gradually align. 
\end{proof}
\textbf{Remark}. Fig.~\ref{fig:verify-eigen-alignment} shows numerical simulation of the symmetrized dynamics (Eqn.~\ref{eq:symmetric-appendix}). If $K(t)$ has negative eigenvalues, then even if $W_p$ and $F$ have already approximately aligned, the dynamics is also unstable and might diverge due to noise and/or numerical instability.

\begin{figure}
    \centering
    \includegraphics[width=0.8\textwidth]{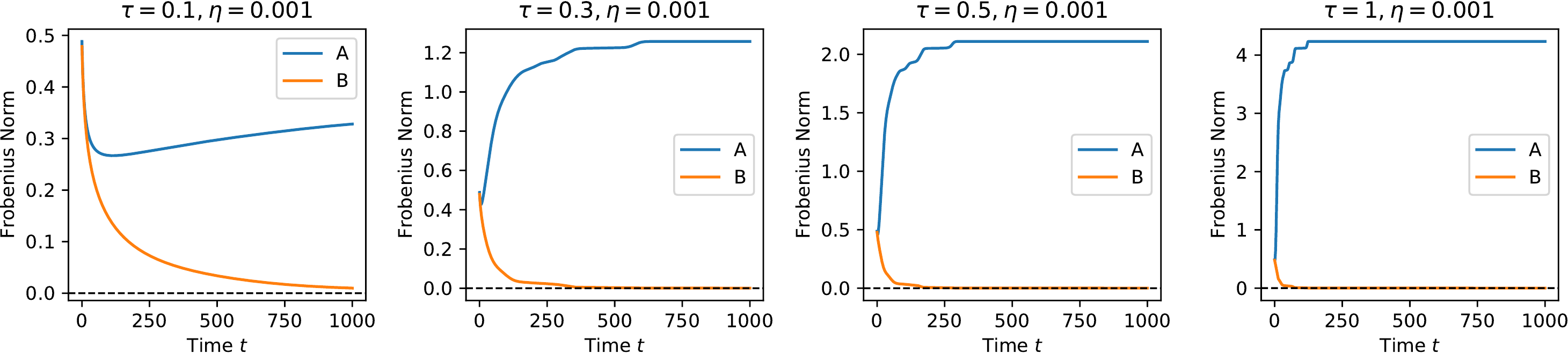}
    \includegraphics[width=0.8\textwidth]{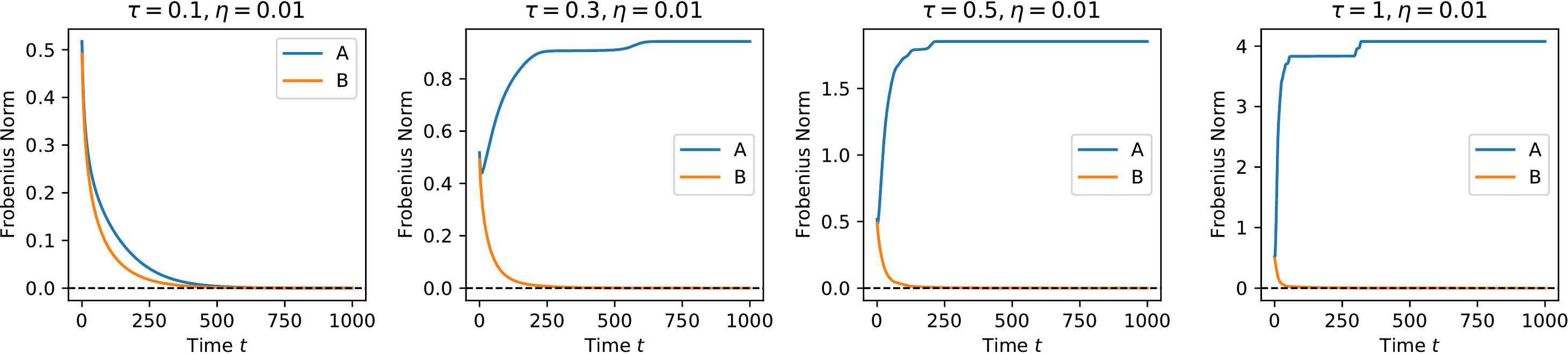}
    \includegraphics[width=0.8\textwidth]{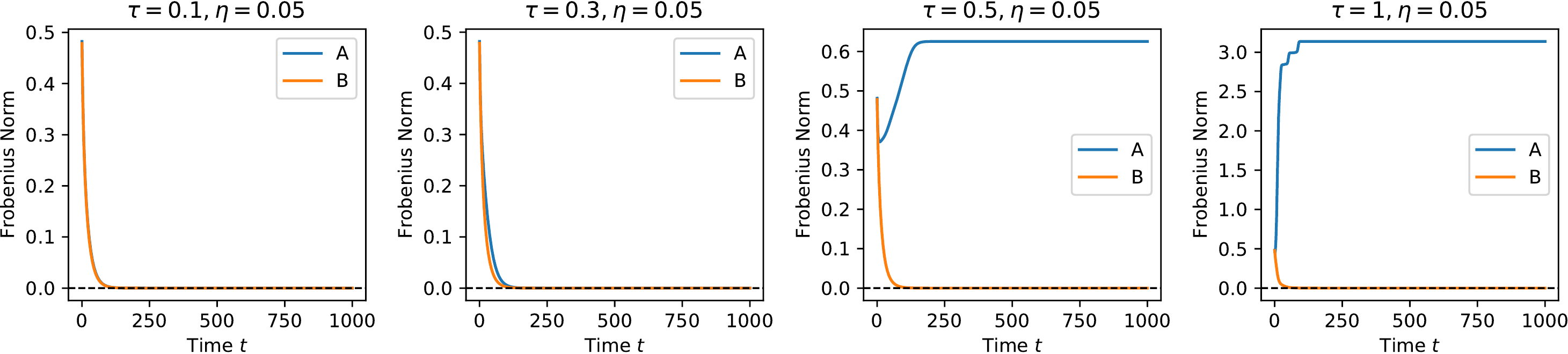}
    \caption{\small Dynamics of the symmetric $A:=(W_p+W_p^\t)/2$ and asymmetric part $B:=(W_p-W_p^\t)/2$ of $W_p$ under different \emph{time-independent} $\tau$ of Eqn.~\ref{eq:wp-assumption-1-2}. Each row is a different weight decay $\eta$ (i.e., $\eta = 0.001$, $0.01$ and $0.05$). When $\eta$ is large and/or $\tau$ is small, $\|A\|_F$ can also be dragged to zero, which is consistent with analysis in Sec.~\ref{sec:decoupled-dynamics} (\underline{Obs\#4} and \underline{Obs\#5}). On the other hand, $\|B\|_F$ always seems to vanish over time. In this numerical simulation, we set $F = W_p^\t W_p$ following invariant in Theorem~\ref{thm:invariance} with $C = 0$.}
    \label{fig:A-B-supp}
\end{figure}

Fig.~\ref{fig:A-B-supp} shows a numerical simulation of Eqn.~\ref{eq:wp-assumption-1-2} (dynamics with Assumption~\ref{assumption:ema} and Assumption~\ref{assumption:isotropic} but without the symmetric dynamics). We can clearly see that the asymmetric component converges to zero. 

\textbf{When eigenspace aligns exactly.} Let $U$ be the common eigenvectors. $W_p = U\Lambda_{W_p} U^\t$ where $\Lambda_{W_p} = \diag[p_1, p_2, \ldots, p_d]$, $F=U\Lambda_F U^\t$ where $\Lambda_F = \diag[s_1, s_2, \ldots, s_d]$.

In this case, the time derivatives $\dot W_p$ and $\dot F$ can all be written as decoupled form: $\dot W_p = U G_1 U^\t$ and $\dot F = U G_2 U^\t$ where $G_1$ and $G_2$ are diagonal matrices. In other words, they are both \emph{decoupled} into each eigen mode, and so does the future value of $W_p$ and $F$. Then $U$ won't change over time. 

To see why, we consider the general case where we have a symmetric matrix $M(t)$ with eigen decomposition $M(t) = U(t)D(t)U^\t(t)$. $M$ follows $\dot M = U(t)G(t) U^\t(t)$ where $G(t)$ is an arbitrary diagonal matrix. 

To see why $\dot U = 0$, at each time step we have:
\begin{equation}
    \dot M = \dot U D U^\t + U\dot D U^\t + UD\dot U^\t = UGU^\t
\end{equation}
since $U$ is unitary, we have:
\begin{equation}
    U^\t \dot U D + D\dot U^\t U = G - \dot D
\end{equation}
Since $U^\t(t) U(t) = I$, we have $\dot U^\t U + U^\t \dot U = 0$ so $Q := U^\t \dot U$ is a skew-symmetric matrix and we have 
\begin{equation}
    QD - DQ = G-\dot D 
\end{equation}
Since the right hand side is a diagonal matrix, checking each entry and we have $q_{ij}d_j - q_{ij}d_i = 0$ for $i\neq j$. If $M$ has distinctive eigenvalues, then we know $q_{ij} = 0$ for $i\neq j$. $Q$ is skew-symmetric so $q_{ii}=0$. So $Q = U^\t \dot U = 0$ and thus $\dot U = 0$. 
If $M$ has duplicated eigenvalues, then we can show $q_{ij} = 0$ for any $d_i \neq d_j$. Within high-dimensional eigenspace for duplicated eigenvalues, its eigen-decomposition is not unique and we can always pick the eigenspace within each duplicated eigenspace so that $\dot U = 0$.  

Therefore, we just multiply $U^\t$ and $U$ to Eqn.~\ref{eq:symmetric-appendix} and the system becomes decoupled. Then after some algebraic manipulation, we arrive at the following: 
\begin{eqnarray}
    \dot p_j &=& \alpha_\pred (1+\sigma^2)s_j\left[\frac{\tau}{1+\sigma^2} - p_j\right] - \eta p_j \label{eq:p-2} \\
    \dot s_j &=& 2(1+\sigma^2)p_js_j\left[\frac{\tau}{1+\sigma^2} -p_j\right] - 2\eta s_j \label{eq:s-2}  
\end{eqnarray}
Multiply Eqn.~\ref{eq:p-2} with $2\alpha^{-1}_\pred p_j$ and subtract with Eqn.~\ref{eq:s-2}, we get:
\begin{equation}
    2\alpha^{-1}_\pred p_j\dot p_j - \dot s_j = -2\eta\alpha^{-1}_\pred p_j^2 + 2\eta s_j
\end{equation}
which gives 
\begin{eqnarray}
    \alpha_\pred^{-1} \left(\frac{\dd p_j^2}{\dd t} + 2\eta p_j^2\right) &=& \dot s_j + 2\eta s_j \\
    \alpha_\pred^{-1} \frac{\dd}{\dd t}(e^{2\eta t}p_j^2) &=& \frac{\dd}{\dd t}(e^{2\eta t}s_j) \\
    \alpha_\pred^{-1} e^{2\eta t}p_j^2 &=& e^{2\eta t}s_j - c_j \\
     \alpha_\pred^{-1} p_j^2(t) &=& s_j(t) - e^{-2\eta t}c_j
\end{eqnarray}
Therefore, we have integral $s_j(t) = \alpha_\pred^{-1} p_j^2(t) + c_j e^{-2\eta t}$. For finite weight decay ($\eta > 0$), we could simply expect $s_j(t) \approx \alpha_\pred^{-1} p_j^2(t)$. 

On the other hand, the dynamics of $\tau$ is:
\begin{equation}
    \dot W_\ema = \beta (W - W_\ema) 
\end{equation}
Applying our assumption about EMA (Assumption~\ref{assumption:ema}) $W_\ema(t) = \tau(t) W(t)$, then we have:
\begin{eqnarray}
    \dot \tau W + \tau \dot W &=& \beta(1-\tau)W \\
    \dot \tau W W^\t + \tau \dot W W^\t &=& \beta(1-\tau)WW^\t \\
    2\dot \tau F + \tau \dot F &=& 2\beta(1-\tau) F
\end{eqnarray}
When $F$ and $W_\pred$ aligns, we have $\dot F$ all in the same eigen space. 
\begin{equation}
    \dot F = -(1+\sigma^2)\{W_\pred^2, F\} + \tau\{W_\pred, F\} - 2\eta F
\end{equation}
So the eigenvectors $U$ won't change and thus we have:
\begin{equation}
    2\dot \tau s_j + \tau \dot s_j = 2\beta(1-\tau) s_j
\end{equation}
or 
\begin{equation}
    \dot \tau = \beta(1-\tau) - \tau \frac{\dot s_j}{2s_j}
\end{equation}
which has a close form solution when $c_j = 0$. Note that in the case, we have $s_j = \alpha_\pred^{-1}p_j^2$ and thus $\dot s_j = 2\alpha^{-1}_\pred p_j\dot p_j$ and we have:
\begin{equation}
    \dot \tau = \beta(1-\tau) - \tau\frac{\dot p_j}{p_j}
\end{equation}
or 
\begin{equation}
    \dot \tau + \tau\left(\frac{\dot p_j}{p_j} + \beta\right) = \beta
\end{equation}
or 
\begin{equation}
    \frac{\dd}{\dd t} (e^{f(t)}\tau) = \beta e^{f(t)}
\end{equation}
where $f(t) = \int (\dot p_j / p_j + \beta) \dd t = \ln p_j + \beta t$ and thus $e^{f(t)} = e^{\beta t} p_j$. Take integral on both side and we have 
(here $\tau(0) = 0$ is the initial condition):
\begin{equation}
    e^{\beta t}p_j \tau = \beta \int_0^t e^{\beta t'} p_j(t')\dd t 
\end{equation}
which is:
\begin{equation}
    \tau_j(t) = p^{-1}_j(t)\beta e^{-\beta t} \int_0^t p_j(t')e^{\beta t'} \dd t
\end{equation}

\subsection{Section~\ref{sec:decoupled-dynamics}}
\textbf{Monotonicity of $p_{j-}^*$ with respect to $\eta$ and $\tau$}. Note that 
\begin{equation}
    p_{j-}^* = \frac{\tau -\sqrt{\tau^2 - 4\eta(1+\sigma^2)}}{2(1+\sigma^2)}
\end{equation}
is the (right) boundary of trivial basin $p < p_{j-}^*$ and determines the size of trivial attractive region towards $p_{j0}^* = 0$. It is dependent on $\eta$ and $\tau$. It is clear that $p_{j-}^*$ is a increasing function of $\eta$. This means that if the weight decay $\eta$ is large, so does trivial region (and more eigenvalues will be trapped to trivial solution).  

On the other hand, we can compute the derivative of $g(x) = x - \sqrt{x^2 - c}$ for $c > 0$ and $x^2 > c$:
\begin{equation}
    \frac{\dd g}{\dd x} = 1 - \frac{1}{\sqrt{1 - c/x^2}} < 0
\end{equation}
So $g(x)$ is a decreasing function with respect to $x$. Or $p^*_{j-}$ is a decreasing function with respect to $\tau$.

\section{Section~\ref{sec:experiment-with-direct-pred}}
\textbf{Experiment setup}. Unless explicitly stated, in all our experiments, we use ResNet-18 as the backbone network, two-layer MLP (with BN and ReLU) as the projector, and a linear predictor. For STL-10 and CIFAR-10, we use SGD as the optimizer with learning rate $\alpha = 0.03$, momentum $0.9$, weight decay $\bar\eta=0.0004$ and EMA parameter $\gamma_\ema=0.996$. The batchsize is 128. Each setting is repeated 5 times to compute mean and standard derivation. We report final number as ``$\mathrm{mean}{\pm}\mathrm{std}$''.

\section{Analysis of BYOL and SimSiam learning dynamics without isotropic assumptions on data}
\label{sec:exactsolbyol}

In the main paper we focused on isotropic data assumptions to obtain analytic insights into when and why BYOL and SimSiam learning dynamics avoid representational collapse.  Here we provide an alternate perspective using a different assumption, involving decoupled initial conditions, that enables us to address the case of learning with non-isotropic data.  First, we recall the data generation and augmentation process.   
Let $\vx$ be a data point drawn from the data distribution $p(\vx)$ and let $\vx_1$ and $\vx_2$ be two augmented views of
$\vx$: $\vx_1, \vx_2\sim p_\aug(\cdot|\vx)$ where $p_\aug(\cdot | \vx)$ is the augmentation distribution.  
Let $\Sigma^s = \ee{\vx_1 \vx_1^\t}$ be the correlation matrix of a single augmented view $\vx_1$ of the data $\vx$, and let 
$\Sigma^d = \ee{\vx_1 \vx_2^\t} $ be the correlation matrix between two augmented views $\vx_1$ and $\vx_2$ of the same data point $\vx$.
In the notation of the main paper, $\Sigma^s$ and $\Sigma^d$ can be decomposed as $\Sigma^s = X + X'$ and $\Sigma^d = X$, where $X = \ee{\bar\vx\bar\vx^\t}$ and $\bar\vx(\vx) := \eee{\vx'\sim p_\aug(\cdot|\vx)}{\vx'}$ is the average augmented view of a data point $\vx$. 
In turn $X' = \eee{\vx}{\var_{\vx'|\vx}[\vx']}$ is the covariance matrix $\var_{\vx'|\vx}[\vx']$ of augmented views $\vx'$ conditioned on $\vx$, subsequently averaged over the data $\vx$.  Intuitively, $X$ is the correlation matrix of augmentation averaged data, while $X'$ is the augmentation covariance matrix averaged over data. 

Also recall that the BYOL learning dynamics, without weight decay, is given by 
\begin{eqnarray}
    \dot W       \!\!&\!\! = \!\!&\!\! W^\t_p \left(- W_\pred W \Sigma^s + W_\ema \Sigma^d \right) \label{eq:onlinelindyn} \\
    \dot W_\pred \!\!&\!\! = \!\!&\!\! \alpha_\pred\left(- W_p W \Sigma^s + W_\ema \Sigma^d \right) W^\t   \label{eq:predlindyn} \\
    \dot W_\ema  \!\!&\!\! = \!\!&\!\! \beta (- W_\ema + W)  \label{eq:targetlindyn}
\end{eqnarray}
SimSiam learning dynamics is a special case in which $W_\ema=W$ and the final equation is ignored. 

We first derive exact fixed point solutions to both BYOL and SimSiam learning dynamics in this setting.  We then discuss specific models for data distributions and augmentation procedures, and show how the fixed point solutions  depend on both data and augmentation distributions.  We then discuss how our theory reveals a fundamental role for the predictor in avoiding collapse in BYOL solutions.  Finally, we derive a highly reduced three dimensional description of BYOL and SimSiam learning dynamics, assuming decopuled initial conditions, that provides considerable insights into dynamical mechanisms enabling both to avoid collapsed solutions without negative pairs to force apart representations of different objects.   

\subsection{The fixed point structure of BYOL and Simsiam learning dynamics.}

Examining \eqref{eq:onlinelindyn}-\eqref{eq:targetlindyn}, we find sufficient conditions for a fixed point given by $W_\pred W \Sigma^s = W_\ema \Sigma^d$ and $W = W_\ema$. Note these are sufficient conditions for fixed points of both BYOL and SimSiam.  Inserting the second equation into the first and right multiplying both sides by $[\Sigma^s]^{-1}$ (assuming $\Sigma^s$ is invertible), yields a manifold of fixed point solutions in $W_1$ and $W_2$ satisfying the nonlinear equation
\begin{equation}
\label{eq:byolfpcond}
    W_\pred W  = W \Sigma^d [ \Sigma^s ]^{-1}.
\end{equation}

This constitutes a set of $n_1 \times n_2$ nonlinear equations in ($n_1 \times n_2) + (n_2 \times n_2$) unknowns, yielding generically a nonlinear manifold of solutions in $W_1$ and $W_2$ of dimensionality $n_2 \times n_2$ corresponding to the number of predictor parameters.  For concreteness, we will assume that $n_2 \leq n_1$, so that the online and target networks perform dimensionality reduction.  Then a special class of solutions to \eqref{eq:byolfpcond} can be obtained by assuming the $n_2$ rows of $W$ correspond to $n_2$ left-eigenvectors of $\Sigma^d [ \Sigma^s ]^{-1}$ and $W_\pred$ is a diagonal matrix with the corresponding eigenvalues.  This special class of solutions can then be generalized by a transformation $W_\pred \rightarrow S W_\pred S^{-1}$ and $W \rightarrow S W$ where $S$ is any invertible $n_2$ by $n_2$ matrix.  Indeed this transformation is a symmetry of \eqref{eq:byolfpcond}, which defines the solution manifold.  In addition to these families of solutions, the collapsed solution $W = W_\pred = W_\ema = 0$ also exists.

\subsection{Illustrative models for data and data augmentation}

The above section suggests that the top eigenmodes of $\Sigma^d [ \Sigma^s ]^{-1}$ control the non-collapsed solutions. Here we make this result more concrete by giving illustrative examples of data distributions and data augmentation procedures, and the resulting properties of $\Sigma^d [ \Sigma^s ]^{-1}$.  

\paragraph{Multiplicative scrambling.} Consider for example a multiplicative subspace scrambling model.
In this model, data augmentation scrambles a subspace by multiplying by a random Gaussian matrix, while identically preserving the orthogonal complement of the subspace.  
In applications, the scrambled subspace could correspond to a space of nuisance features, while the preserved subspace could correspond to semantically important features.  Indeed many augmentation procedures, including random color distortions and blurs, largely preserve important semantic information, like object identity in images. 

More precisely, we consider a random scrambling operator ${A}$ which only scrambles data vectors $\vx$ within a fixed $k$ dimensional subspace spanned by the orthonormal columns of the $n_0 \times k$ matrix ${U}$. 
Within this subspace, data vectors are scrambled by a random Gaussian $k \times k$ matrix ${B}$. 
Thus ${A}$ takes the form ${A} = {P}^c + UBU^T$ where ${P}^c = {I} - {UU}^T$ is a projection operator onto the $n_0-k$ dimensional conserved, semantically important, subspace orthogonal to the span of the columns of ${U}$, and the elements of ${B}$ are i.i.d. zero mean unit variance Gaussian random variables so that $\eee{}{B_{ij} B_{kl}} = \delta_{ik} \delta_{jl}$. 
Under this simple model, the augmentation average $\bar\vx(\vx) := \eee{\vx'\sim p_\aug(\cdot|\vx)}{\vx'}$ becomes $\bar\vx(\vx) = P^c \vx$. Thus, intuitively, under multiplicative subspace scrambling, the only aspect of a data vector that survives averaging over augmentations is the projection of this data vector onto the preserved subspace. 
Then the correlation matrix of two different augmented views is $\Sigma^d = P^c \Sigma^x P^c$ while the correlation matrix of two identical views is $\Sigma^s = \Sigma^x$ where $\Sigma^x \equiv \eee{\vx \sim p(\cdot)}{\vx \vx^T}$ is the correlation matrix of the data distribution.  Thus non-collapsed solutions of both BYOL and SimSiam can correspond to principal eigenmodes of  $\Sigma^d [ \Sigma^s ]^{-1} = P^c \Sigma^x P_c [ \Sigma^x ]^{-1}$.  In the special case in which $P^c$ commutes with $\Sigma^x$, we have the simple result that $\Sigma^d [ \Sigma^s ]^{-1} = P^c$, which is completely independent of the data correlation matrix $\Sigma^x$.  Thus in this simple setting BYOL and SimSiam can learn the subspace of features that are identically conserved under data augmentation, independent of how much data variance there is in the different dimensions of this conserved subspace.       

\paragraph{Additive scrambling.} We also consider, as an illustrative example, data augmentation procedures which simply add Gaussian noise with a prescribed noise covariance matrix $\Sigma^n$.  Under this model, we have $\Sigma^s = \Sigma^x + \Sigma^n$ while $\Sigma^d = \Sigma^x$.  Thus in this setting, BYOL learns principal eigenmodes of  $\Sigma^d [ \Sigma^s ]^{-1} =  \Sigma^x [\Sigma^x + \Sigma^n]^{-1}$. Thus intuitively, dimensions with larger noise variance are attenuated in learned BYOL representations. On the otherhand, correlations in the data that are not attenuated by noise are preferentially learned, but the degree to which they are learned is not strongly influenced by the magnitude of the data correlation (i.e. consider dimensions that lie along small eigenvalues of $\Sigma^n$). Note that in the main paper we focused on the case where $\Sigma^x = I$ and $\Sigma^n = \sigma^2 I$.

\subsection{The importance of the predictor in BYOL and SimSiam.}
Here we note that our theory explains why the predictor plays a crucial role in BYOL and SimSiam learning in this simple setting, as is observed empirically in more complex settings.  To see this, we can model the removal of the predictor by simply setting $W_\pred = I$ in all the above equations.  The fixed point solutions then obey  $W = W \Sigma^d [ \Sigma^s ]^{-1}$.  This will only have nontrivial, non-collapsed solutions if $\Sigma^d [ \Sigma^s ]^{-1}$ has eigenvectors with eigenvalue $1$. Rows of $W$ consisting of linear combinations of these eigenvectors will then constitute non-collapsed solutions.    

This constraint of eigenvalue $1$ yields a much more restrictive condition on data distributions and augmentation procedures for BYOL and Simsiam to have non-collapsed solutions.  It can however be satisfied in multiplicative scrambling if an eigenvector of the data matrix $\Sigma^x$ lies in the column space of the projection operator $P^c$ (in which case it is an eigenvector of eigenvalue $1$ of $\Sigma^d [ \Sigma^s ]^{-1} = P^c \Sigma^x P_c [ \Sigma^x ]^{-1}$.  This condition cannot however be generically satisfied for additive scrambling case, in which generically all the eigenvalues of $\Sigma^d [ \Sigma^s ]^{-1} =  \Sigma^x [\Sigma^x + \Sigma^n]^{-1}$ are less than $1$. In this case, without a predictor, it can be checked that the collapsed solution $W = W_\ema = 0$ is stable.

Thus overall, in this simple setting, our theory provides conceptual insight into how the introduction of a predictor is crucial for creating new non-collapsed solutions for both BYOL and SimSiam, even though the  predictor confers no new expressive capacity in allowing the online network to match the target network. 

\subsection{Reduction of BYOL learning dynamics to low dimensions}

The full learning dynamics in \eqref{eq:onlinelindyn} to \eqref{eq:targetlindyn} constitutes a set of high dimensional nonlinear ODEs which are difficult to solve from arbitrary initial conditions. However, there is a special class of {\it decoupled} initial conditions which permits additional insight.  Consider the special case in which  $\Sigma^s$ and $\Sigma^d$ commute, and so are simultaneously diagonalizable and share a common set of eigenvectors, which we denote by $\vu^\alpha \in \rr^{n_0}$.  Consider also a special set of initial conditions where each row of $W$ and the corresponding row of $W_\ema$ are both proportional to one of the eigenmodes $\vu^\alpha$, with scalar proportionality constants $w^\alpha$ and $w_\ema^\alpha$ respectively, and $W_\pred$ is diagonal, with the corresponding diagonal element given by $w_\pred^\alpha$.  Then it is straightforward to see that under the dynamics  in \eqref{eq:onlinelindyn} to \eqref{eq:targetlindyn}, that the structure of this initial condition will remain the same, with only the scalars $w^\alpha$,  $w_\ema^\alpha$ and $w_\pred^\alpha$ changing over time.  Moreover, the scalars decouple across the different indices $\alpha$, and the dynamics are driven by the eigenvalues $\lambda_s^\alpha$ and $\lambda_d^\alpha$ of $\Sigma_s$ and $\Sigma_d$ respectively.  Inserting this special class of initial conditions into the dynamics in \eqref{eq:onlinelindyn} to \eqref{eq:targetlindyn}, and dropping the $\alpha$ index, we find the dynamics of the triplet of scalars is given by
\begin{align}
    \label{eq:onlinelindyns}
    \frac{d w_\pred}{dt} & = \alpha_p \left[ w_\ema \lambda_d - w_\pred w \lambda_s \right] w \\
    \label{eq:predlindyns}
    \frac{d w}{dt} & =  w_\pred \left[ w_\ema \lambda_d - w_\pred w \lambda_s \right] \\
    \label{eq:targetlindyns}
    \frac{dw_\ema}{dt} & = \beta(-w_\ema + w).
\end{align}
Alternatively, this low dimensional dynamics can be obtained from \eqref{eq:onlinelindyn} to \eqref{eq:targetlindyn} not only by considering a special class of decoupled initial conditions, but also by considering the special case where every matrix is simply a $1$ by $1$ matrix, making the scalar replacements $W \rightarrow w$, $W_\pred \rightarrow w_\pred$, $W_\ema \rightarrow w_\ema$, $\Sigma^s \rightarrow \lambda_s$, and $\Sigma^d \rightarrow \lambda_d$.  Note furthermore that this $3$ dimensional dynamical system is equivalent to that studied in the main paper under the change of variables $s=w^2$ and $\tau = w_\ema / w$ and the special case of $\lambda_s = 1+\sigma^2$ and $\lambda_d = 1$.   

\begin{figure}
    \centering
    \includegraphics[width=\textwidth]{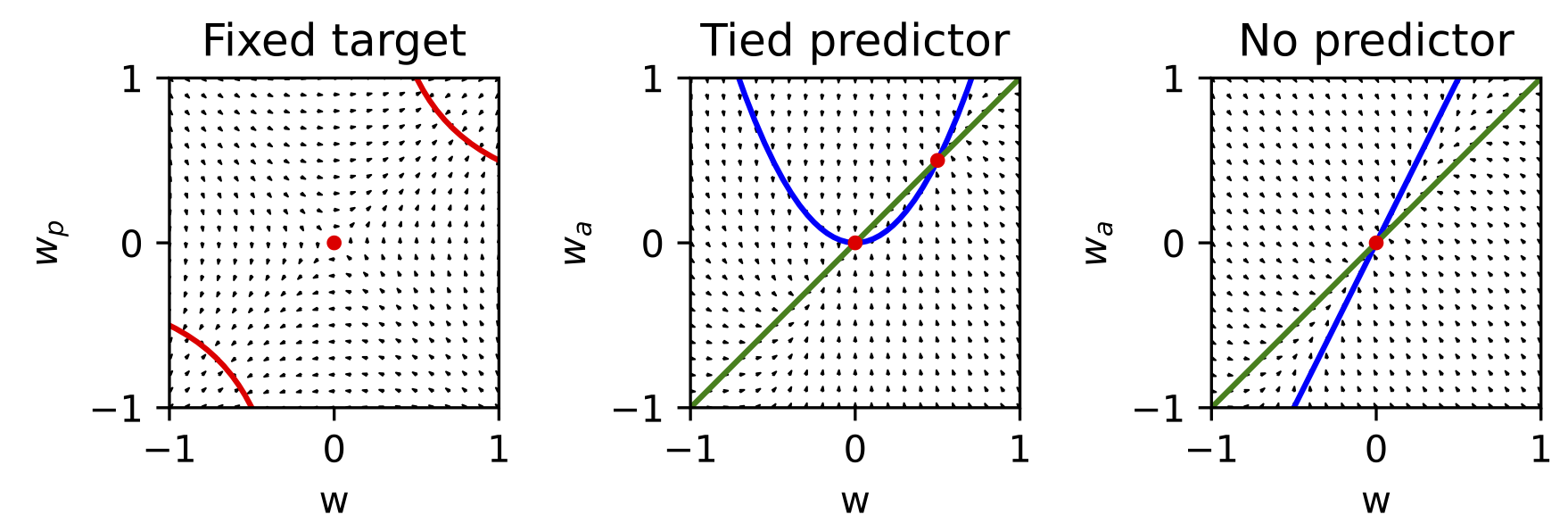}
    \vspace{-0.1in}
    \caption{\small A visualization of BYOL dynamics in low dimensions. \textbf{Left}: Black arrows denote the vector field of the flow in the $w$ and $w_\pred$ plane of online and predictor weights in Eqns. \ref{eq:onlinelindyns} and \ref{eq:predlindyns} when the target network weight $w_\ema$ is fixed to $1$.  For all $3$ panels, $\lambda_s=1$, $\lambda_d=1/2$, and $\alpha_\pred = \beta = 1$. All flow field vectors are normalized to unit length to indicate direction of flow alone.  The red curve shows the hyperoblic manifold of stable fixed points $w_\pred w = w_\ema \lambda_d \lambda_s^{-1}$, while the red point at the origin is an unstable fixed point. For a fixed target network, the online and predictor weights will cooperatively amplify each other to escape the collapsed solution at the origin. \textbf{Middle}: A visualization of the full low dimensional BYOL dynamics in Eqns \ref{eq:onlinelindyns}-\ref{eq:targetlindyns} when the online and predictor weights are tied so that $w = w_\pred$.  The green curve shows the nullcline $w_\ema = w$ corresponding to $\frac{dw_\ema}{dt}=0$ and the blue curve shows part of the nullcline $\frac{dw}{dt}=0$ corresponding to $w^2 = w_\ema \lambda_d \lambda_s^{-1}$.  The intersection of these two nullclines yields two fixed points (red dots): an unstable collapsed solution at the origin $w=w_\ema=0$, and a stable non-collapsed solution with $w_\ema=w$ and $w = \lambda_d \lambda_s^{-1}$. \textbf{Right}: A visualization of dynamics in Eqns \ref{eq:onlinelindyns}-\ref{eq:targetlindyns} when the the predictor is removed, so that $w_2$ is fixed to $1$.  The resulting two dimensional flow field on $w$ and $w_\ema$ is shown (black arrows).  The green curve shows the nullcline $w=w_\ema$ corresponding to $\frac{dw_\ema}{dt}=0$, while the blue curve shows the nullcline $w = w_\ema \lambda_d \lambda_s^{-1}$.  The slope of this nullcline is $\lambda_s \lambda_d^{-1} > 1$.  The resulting nullcline structure yields a single fixed point at the origin which is stable.  Thus there only exists a collapsed solution. In the special case where $\lambda_s \lambda_d^{-1} = 1$, the two nullclines coincide, yielding a one dimensional manifold of solutions.}   
    \vspace{-0.1in}
    \label{fig:lowdimbyol}
\end{figure}

The fixed point conditions of this dynamics are given by $w_\ema = w$ and $w_\pred w = w_\ema \lambda_d \lambda_s^{-1}$.  Thus the collapsed point $w = w_\pred = w_\ema = 0$ is a solution.  Additionally $w_\pred = \lambda_d \lambda_s^{-1}$ and $w=w_\ema$ taking any value is also a family of non-collapsed solutions. We can understand the three dimensional dynamics intuitively as follows when $\beta$ is much less than both $1$ and $\alpha_p$, so that the dynamics of $w$ and $w_\pred$ are very fast relative to the dynamics of $w_\ema$.  In this case, the target network evolves very slowly compared to the online network, as is done in practice. For simplicity we use the same learning rate for the predictor as we do for the online network (i.e. $\alpha_p=1$).  In this situation, we can treat $w_\ema$ as approximately constant on the fast time scale over which the online and predictor weights $w$ and $w_\pred$ evolve. Then the joint dynamics in \eqref{eq:onlinelindyns} and \eqref{eq:predlindyns} obeys gradient descent on the error function
\begin{equation}
E = \frac{\lambda_s}{2}( w_\ema \lambda_d \lambda_s^{-1} - w_\pred w)^2.
\end{equation}
Iso-contours of constant error are hyperbolas in the $w$ by $w_\pred$ plane, and for fixed $w_\ema$, the origin $w = w_\pred = 0$ is a saddle point, yielding an unstable fixed point (see Fig.~\ref{fig:lowdimbyol} (left)).  From generic initial conditions, $w$ and $w_p$ will then cooperatively amplify each other to rapidly escape the collapsed solution at the origin, and approach the zero error hyperbolic contour  $w_\pred w = w_\ema \lambda_d \lambda_s^{-1}$ where $w_\ema$ is close to its initial value.  Then the slower target network $w_\ema$ will adjust, slowly moving this contour until $w_\ema = w$.  The more rapid dynamics of $w$ and $w_\pred$ will hug the moving contour $w_\pred w = w_\ema \lambda_d \lambda_s^{-1}$ as $w_\ema$ slowly adjusts. In this fashion, the joint fast dynamics of $w$ and $w_\pred$, combined with the slow dynamics of $w_\ema$, leads to a nonzero fixed point for all $3$ values, despite the existence of a collapsed fixed point at the origin.  Moreover, the larger the ratio $\lambda_d \lambda_s^{-1}$, which is determined by the data and augmentation, the larger the final values of both $w$ and $w_\pred$ will tend to be. 

We can obtain further insight by noting that the submanifold $w = w_\pred$, in which the online and predictor weights are tied, constitutes an invariant submanifold of the dynamics in Eqns. \ref{eq:onlinelindyns} to \ref{eq:targetlindyns}; if $w=w_\pred$ at any instant of time, then this condition holds for all future time.  Therefore we can both analyze and visualize the dynamics on this two dimensional invariant submanifold, with coordinates $w=w_\pred$ and $w_\ema$ (Fig.~\ref{fig:lowdimbyol} (middle)).  This analysis clearly shows an unstable collapsed solution at the origin, with $w=w_\ema=0$, and a stable non-collapsed solution at $w=w_\ema=\lambda_d \lambda_s^{-1}$. 

We note again, that the generic existence of these non-collapsed solutions in Fig.~\ref{fig:lowdimbyol} depends critically on the presence of a predictor with adjustable weights $w_\pred$.  Removing the predictor corresponds to forcing $w_\pred=1$, and non-collapsed solutions cannot exist unless $\lambda_d = \lambda_s$, as demonstrated in Fig.~\ref{fig:lowdimbyol} (right).  Thus, remarkably, in BYOL in this simple setting, the introduction of a predictor network plays a crucial role, even though it neither adds to the expressive capacity of the online network, nor improves its ability to match the target network.  Instead, it plays a crucial role by dramatically modifying the learning dynamics (compare e.g. Fig~\ref{fig:lowdimbyol} middle and right panels), thereby enabling convergence to noncollapsed solutions through a dynamical mechanism whereby the online and predictor network cooperatively amplify each others' weights to escape collapsed solutions ( Fig.~\ref{fig:lowdimbyol} (left)). 

Overall, this analysis of BYOL learning dynamics provides considerable insight into the dynamical mechanisms enabling BYOL to avoid collapsed solutions, without negative pairs to force apart representations, in what is likely to be the simplest nontrivial setting.  Further analysis on this model, in direct analogy to the analysis performed on the equivalent $3$ dynamical system (derived under different assumptions) studied in the main paper, can yield similar insights into the dynamics of BYOL and SimSiam under various conditions on learning rates.   

%\bibliographystyle{icml2021}
%\bibliography{references}

\end{document}